%% file: main.tex
\documentclass[runningheads]{llncs}
\usepackage{graphicx}
\usepackage{comment}
\usepackage{amsmath,amssymb} 
\usepackage{color}
\usepackage{booktabs}
\usepackage{balance}
\usepackage{lscape}
\usepackage{tabularx}
\usepackage{arydshln}
\usepackage{multirow}
\usepackage{array}
\usepackage{arydshln}
\usepackage{subcaption}

\begin{document}
\pagestyle{headings}
\mainmatter
\def\ECCVSubNumber{2783}  

\title{Captioning Images Taken by \\People Who Are Blind}

\titlerunning{Captioning Images Taken by People Who Are Blind}
\author{Danna Gurari, Yinan Zhao, Meng Zhang, Nilavra Bhattacharya}
\authorrunning{D. Gurari et al.}
\institute{University of Texas at Austin}
\maketitle

\input{abstract}
\input{introduction}
\input{related-works}
\input{dataset-creation}
\input{dataset-analysis}
\input{algorithm-benchmarking}
\input{conclusions}

\noindent
\paragraph{\textbf{Acknowledgements.}}
We thank Meredith Ringel Morris, Ed Cutrell, Neel Joshi, Besmira Nushi, and Kenneth R. Fleischmann for their valuable discussions about this work.  We thank Peter Anderson and Harsh Agrawal for sharing their code for setting up the EvalAI evaluation server.  We thank the anonymous crowdworkers for providing the annotations.  This work is supported by National Science Foundation funding (IIS-1755593), gifts from Microsoft, and gifts from Amazon.

\clearpage
{\small
\bibliographystyle{ieee}
\bibliography{myReferences}
}

\clearpage
\input{supplementary-materials.tex}

\end{document}

%% file: abstract.tex
\begin{abstract}
While an important problem in the vision community is to design algorithms that can automatically caption images, few publicly-available datasets for algorithm development directly address the interests of real users.  Observing that people who are blind have relied on (human-based) image captioning services to learn about images they take for nearly a decade, we introduce the first image captioning dataset to represent this real use case.  This new dataset, which we call VizWiz-Captions, consists of over 39,000 images originating from people who are blind that are each paired with five captions.  We analyze this dataset to (1) characterize the typical captions, (2) characterize the diversity of content found in the images, and (3) compare its content to that found in eight popular vision datasets.  We also analyze modern image captioning algorithms to identify what makes this new dataset challenging for the vision community.  We publicly-share the dataset with captioning challenge instructions at \texttt{https://vizwiz.org}.

\end{abstract}

%% file: introduction.tex
\section{Introduction}
A popular computer vision goal is to create algorithms that can replicate a human's ability to caption any image~\cite{bai2018surveyautomaticimage,hossain2019comprehensivesurveydeep,srivastava2018surveyautomaticimage}.  Presently, we are witnessing an exciting transition where this dream of automated captioning is advancing into a reality, with automated image captioning now a feature available in several popular technology services.  For example, companies such as Facebook and Microsoft are providing automated captioning in their social media~\cite{Howdoesautomatic} and productivity (e.g., Power Point)~\cite{Addalternativetext} applications to enable people who are blind to make some sense of images they encounter in these digital environments. 

\begin{figure}[t!]
\centering
\includegraphics[width=\textwidth]{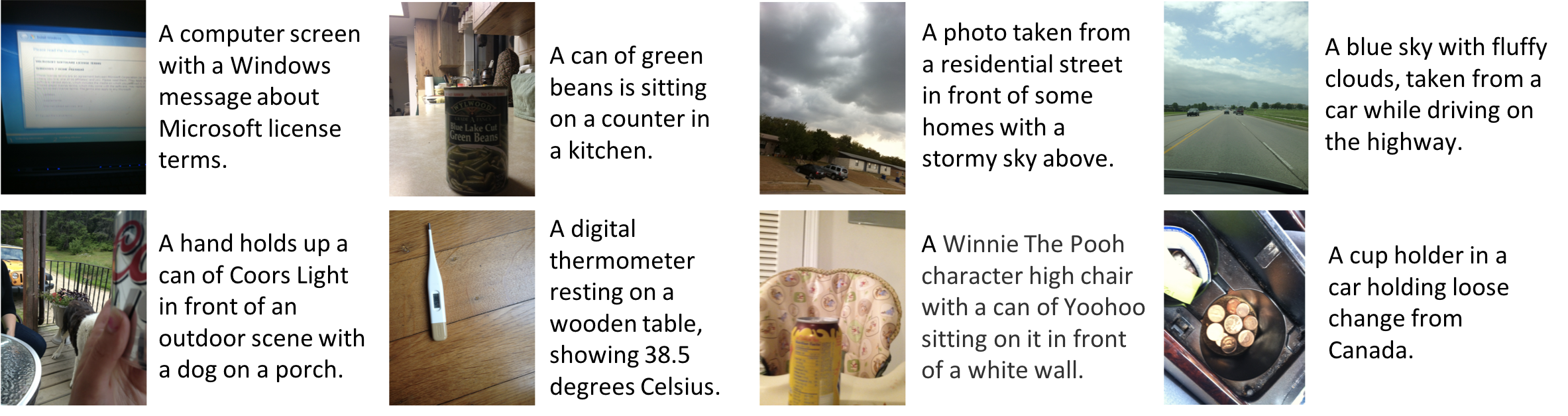}
\caption{Examples of captioned images in our new dataset, which we call VizWiz-Captions.  These exemplify that images often contain text, exhibit a high variability in image quality, and contain a large diversity of content.}
\label{fig_exampleCaptions}
\end{figure}

While much of the progress has been fueled by the recent creation of large-scale, publicly-available datasets (needed to train and evaluate algorithms), a limitation is that most existing datasets were created in contrived settings.  Typically, crowdsourced workers were employed to produce captions for images curated from online, public image databases such as Flickr~\cite{agrawal2018nocapsnovelobject,chen2015MicrosoftCOCOcaptions,gan2017StylenetGeneratingattractive,harwath2015Deepmultimodalsemantic,hodosh2013Framingimagedescription,krishna2017VisualgenomeConnecting,yoshikawa2017StaircaptionsConstructing,young2014imagedescriptionsvisual}.  Yet, we have observed over the past decade that people have been collecting image captions to meet their real needs.  Specifically, people who are blind have sought descriptions\footnote{Throughout, we use ``caption" and ``description" interchangeably.} from human-powered services~\cite{BeSpecular,bigham2010VizWiznearlyrealtime,salisbury2017ScalableSocialAlt,vonahn2006Improvingaccessibilityweb,zhong2015RegionspeakQuickcomprehensive} to learn more about pictures they take of their visual surroundings.  Unfortunately, images taken by these real users in the wild often exhibit dramatically different conditions than observed in the contrived environments used to design modern algorithms, as we will expand upon in this paper.  Examples of some of the unique characteristics of images taken by real users of image captioning services are exemplified in Figure~\ref{fig_exampleCaptions}.  The consequence is that algorithms tend to perform poorly when deployed on their images.

To address the above problem, we introduce the first publicly-available captioning dataset that consists of images taken by people who are blind.  This dataset builds off of prior work which supported real users of a mobile phone application to submit a picture and, optionally, record a spoken question in order to learn about their images~\cite{bigham2010VizWiznearlyrealtime}.  We crowdsourced captions for 39,181 images that were submitted.  We also collected metadata for each image that indicates whether text is present and the severity of image quality issues to enable a systematic analysis around these factors.  We call this dataset VizWiz-Captions.  

We then characterize how our new dataset relates to the momentum of the broader vision community.  To do so, we characterize how the captioned content relates/differs to what is contained in eight popular vision datasets that support the image captioning, visual question answering, and image classification tasks.  We observe both that VizWiz-Captions shows many distinct visual concepts from those in existing datasets and regularly provides the answers to people's visual questions (Section~\ref{sec:dataset-analysis}).  We also benchmark modern captioning algorithms, and find that they struggle to caption lower quality images.  

We offer this work as a valuable foundation for designing more generalized computer vision algorithms that meet the large diversity of needs for real end users.  Our dataset can facilitate and motivate progress for a broader number of scenarios that face similar complexities.  For example, wearable lifelogging devices, autonomous vehicles, and robots also can result in varying image quality and many images showing textual information (e.g., street signs, billboards) as important real-world challenges that must be handled to solve downstream tasks.  

To facilitate and encourage progress, we organized a dataset challenge and associated workshop to track progress and stimulate discussion about current research and application issues.  Details about the dataset, challenge, and workshop can be found at the following link: \texttt{https://vizwiz.org}.

%% file: related-works.tex
\section{Related Work}
\emph{Captioning Images for People Who are Blind.}
Given the clear wish from people who are blind to receive descriptions of images~\cite{HomeAiraAira,bennett2018HowTeensVisual,bigham2010VizWiznearlyrealtime,burton2012Crowdsourcingsubjectivefashion,macleod2017UnderstandingBlindPeople,petrie2005Describingimagesweb,voykinska2016Howblindpeople}, many human-in-the-loop~\cite{BeSpecular,HomeAiraAira,TapTapSeeBlindVisually,morris2018RichRepresentationsVisual,salisbury2017ScalableSocialAlt,vonahn2006Improvingaccessibilityweb} and automated services~\cite{Addalternativetext,Howdoesautomatic} have emerged to do so.  A challenge shared across such services is what content to describe.  Although there remains a lack of guidance for images taken by people who are blind~\cite{stangl2020person}, it is known that many people who are blind report a preference to receive descriptions of images over nothing (even if inaccurate)~\cite{guinness2018CaptionCrawlerEnabling,salisbury2017ScalableSocialAlt,salisbury2018EvaluatingComplementingVisiontoLanguage,wu2017AutomaticAlttextComputergenerated}.  Accordingly, to facilitate progress on automated solutions for captioning images taken by this population, we introduce a new dataset to represent this use case.  In doing so, we aim to support the design of a cheaper, faster, and more private alternative than is possible with human-based captioning services. 

\emph{Image Captioning Datasets.}
Over the past decade, nearly 20 publicly-shared captioning datasets have been created to support the development of automated captioning algorithms~\cite{agrawal2018nocapsnovelobject,chen2015Dejaimagecaptionscorpus,chen2015MicrosoftCOCOcaptions,elliott2013Imagedescriptionusing,farhadi2010Everypicturetells,gan2017StylenetGeneratingattractive,harwath2015Deepmultimodalsemantic,havard2017Speechcoco600kvisually,hodosh2013Framingimagedescription,kong2014Whatareyou,krishna2017VisualgenomeConnecting,rashtchian2010Collectingimageannotations,shuster2018Engagingimagecaptioning,yoshikawa2017StaircaptionsConstructing,young2014imagedescriptionsvisual,zitnick2013Learningvisualinterpretation}.  The trend has been to include a larger number of examples, relying on scraping images from the web (typically Flickr) to support the growth from a few thousand~\cite{farhadi2010Everypicturetells,feng2008Automaticimageannotation,rashtchian2010Collectingimageannotations} to hundreds of thousands~\cite{chen2015MicrosoftCOCOcaptions,havard2017Speechcoco600kvisually,krishna2017VisualgenomeConnecting} of captioned images in such datasets.  In doing so, such work has strayed from focusing on real use cases.  To help align the vision community to focus on addressing the real interests of people who need image captions, we instead focus on introducing a captioning dataset that emerges from a natural use case.  

Accordingly, our work more closely aligns with the earlier datasets that emerged from authentic image captioning scenarios.  This includes captioned images in newspaper articles~\cite{feng2008Automaticimageannotation} and provided by tour guides about photographs of tourist locations~\cite{grubinger2006iaprtc12benchmark}.  Unlike these prior works, we focus on a distinct use case (i.e., captioning blind photographers' images) and our new dataset is considerably larger (i.e., contains nearly 40,000 images versus 3,361~\cite{feng2008Automaticimageannotation} and 20,000~\cite{grubinger2006iaprtc12benchmark}).  

More generally, to our knowledge, our new captioning dataset is the first that comes with metadata indicating for each image whether text is present and the severity of image quality issues, thereby enabling systematic analysis around these factors.  We expect this new dataset will contribute to the design of more robust, general-purpose captioning algorithms. 

\emph{Content in Vision Datasets.}
The typical trend for curating images for popular vision datasets is to scrape various web search engines for pre-defined categories/search terms.  For example, this is how popular object recognition datasets (e.g., ImageNet~\cite{russakovsky2015Imagenetlargescale} and COCO~\cite{lin2014MicrosoftcocoCommon}), scene recognition datasets (e.g., SUN~\cite{xiao2010SundatabaseLargescale} and Places205~\cite{zhou2014Learningdeepfeatures}), and attribute recognition datasets (e.g., SUN-attributes~\cite{patterson2016CocoattributesAttributes} and COCO-attributes~\cite{patterson2012Sunattributedatabase}) were created.  Observing that automated methods rely on such large-scale datasets to guide what concepts they learn, a question emerges of how well the content in such contrived datasets reflect the interests of real users of image descriptions services.  We conduct comparisons between popular vision datasets and our new dataset to provide such insight.  This analysis is valuable both for highlighting the value of existing datasets to support a real use case and revealing how vision datasets can be improved.

%% file: dataset-creation.tex
\section{VizWiz-Captions}
We now introduce VizWiz-Captions, a dataset that consists of descriptions about images taken by people who are blind.  Our work builds upon two existing datasets that contain images taken by real users of a visual description service~\cite{gurari2019VizWizPrivDatasetRecognizing,gurari2018VizWizGrandChallenge}.  The images in these datasets originate from users of the mobile phone application VizWiz~\cite{bigham2010VizWiznearlyrealtime}, who each submitted a picture with, optionally, a recorded spoken question in order to receive a description of the image or answer to the question (when one was asked) from remote humans.  In total, we used the 39,181 images that are publicly-shared and were not corrupted to obfuscate private content.  Of these, 16\% (i.e., 6,339) lack a question.  We detail below our creation and analysis of this dataset.

\subsection{Dataset Creation}

\paragraph{Image Captioning System.}
To collect captions, we designed our captioning task for use in the crowdsourcing platform Amazon Mechanical Turk (AMT).  To our knowledge, the only public precedent for crowdsourcing image descriptions from crowdworkers for images taken by people who are blind is the VizWiz mobile phone application~\cite{bigham2010VizWiznearlyrealtime}.  This system offered vague instructions to `describe the image'.  Given this vague precedence, we chose to adapt the more concrete task design from the vision community, as described below.

We employed the basic task interface design used by prior work in the vision community~\cite{agrawal2018nocapsnovelobject,chen2015MicrosoftCOCOcaptions,hodosh2013Framingimagedescription,young2014imagedescriptionsvisual}.  It displays the image on the left, instructions on the right, and a text entry box below the instructions for entering the description.  The instructions specify to include at least eight words as well as what not to do when creating the caption (e.g., do not speculate what people in the image might be saying/thinking or what may have happened in the future/past).  

We further augmented the task interface to tailor it to unique characteristics of our captioning problem.  These augmentations resulted both from consultation with accessibility experts and iterative refinement over four pilot studies.  First, to encourage crowdworkers to address the interests of the target audience, we added the instruction to ``Describe all parts of the image that may be important to a person who is blind."  Second, to encourage crowdworkers to focus on the content the photographer likely was trying to capture rather than any symptoms of low quality images that inadvertently arise for blind photographers, we instructed crowdworkers ``DO NOT describe the image quality issues."  However, given that some images could be insufficient quality for captioning, we provided a button that the crowdworker could click in order to populate the description with pre-canned text that indicates this occurred (i.e., ``Quality issues are too severe to recognize visual content.").  Next, to discourage crowdworkers from performing the optical character recognition problem when text is present, we added the following instruction: ``If text is in the image, and is important, then you can summarize what it says. DO NOT use all the specific phrases that you see in the image as your description of the image."  Finally, to enrich our analysis, we asked crowdworkers to provide extra information about each image regarding whether text is present.  

\paragraph{Caption Collection and Post-Processing.}
For each of the 39,181 images, we collected redundant results from the crowd.  In particular, we employed five AMT crowdworkers to complete our task for every image.  We applied a number of quality control methods to mitigate concerns about the quality of the crowdsourced results, summarized in the Supplementary Materials.  In total, we collected 195,905 captions.  All this work was completed by 1,623 crowdworkers who contributed a total of 3,736 person-hours.  With it being completed over a duration of 101.52 hours, this translates to roughly 37 person-hours of work completed every hour.  We post-processed each caption by applying a spell-checker to detect and fix misspelled words.  

%% file: dataset-analysis.tex
\subsection{Dataset Analysis}
\label{sec:dataset-analysis}

\paragraph{Quality of Images.}  
We first examined the extent to which the images were deemed to be insufficient quality to caption.  This is important to check, since people who are blind cannot verify the quality of the images they take, and it is known their images can be poor quality due to improper lighting (i.e., mostly white or mostly black), focus, and more~\cite{brady2013Visualchallengeseveryday,chiu2020assessing,gurari2018VizWizGrandChallenge}.  To do so, we tallied how many of the five crowdworkers captioned each image with the pre-canned text indicating insufficient quality for captioning (i.e., ``Quality issues are too severe...").  The distribution of images for which none to all 5 crowdworkers used this pre-canned text is as follows: 68.5\% for none, 16.7\% for 1 person, 5.9\% for 2 people, 3.6\% for 3 people, 3.1\% for 4 people, and 2.2\% for all 5 people.

We found that the vast majority of images taken by blind photographers were deemed good enough quality that the content can be recognized.  Only 9\% of the images were deemed insufficient quality for captioning by the majority of the crowdworkers.  A further 22.6\% of images were deemed insufficient quality by a minority of the crowdworkers (i.e., 1 or 2).  Altogether, these findings highlight a range of difficulty for captioning, based on the extent to which crowdworkers agreed the images are (in)sufficient quality to generate a caption.  In Section~\ref{sec_algBenchmarking}, we report the ease/difficulty for algorithms to caption images based on this range of perceived difficulty by humans.

\begin{table*}[t!]
  \centering
        \begin{tabular}{ l  c  c  c  c  c  c  c  c  c  c}
    \toprule
       & \multicolumn{5}{c}{\bf Average Count Per Image } & \multicolumn{5}{c}{{\bf Unique Count for All Images}} \\
     \cmidrule(r){2-6} \cmidrule(r){7-11}
       & words & nouns & verbs & adj & spa-rel & words & nouns & verbs & adj & spa-rel \\
    \midrule
       \textbf{Ours} & 13.0 & 4.4 & 0.9 & 1.4 & 1.9 & 24,422 & 16,400 &  4,040 & 8,755 & 275 \\ 
       \cdashline{1-11}
       \textbf{Ours-WithQues} &  13.0 & 4.4 & 0.9 & 1.4 & 1.9 & 22,261 & 14,933 & 3,719 & 7,882 & 244 \\ 
       \textbf{Ours-NoQues} &  13.0 & 4.4 & 0.9 & 1.5 & 1.9 & 10,651 & 7,249 & 1,616 & 3,212 & 120 \\ 
       \cdashline{1-11}
       \textbf{Ours-WithText} &  12.9 & 4.5 & 0.9 & 1.4 & 1.9 & 21,161 & 14,277 & 3,294 & 7,263 & 243  \\ 
       \textbf{Ours-NoText} & 13.1 & 4.2 & 0.9 & 1.6 & 1.9 & 10,711 & 7,114 & 1,933 & 3,508 & 127  \\ 
       \cdashline{1-11}
       \textbf{\cite{chen2015MicrosoftCOCOcaptions}-All} & 11.3 & 3.7 & 1.0 & 0.9 & 1.7 & 30,122 & 19,998 &  6,697 & 9,651 & 381  \\ 
       \textbf{\cite{chen2015MicrosoftCOCOcaptions}-Sample} & 11.3 & 3.7 & 1.0 & 0.9 & 1.7 & 16,966 & 11,211 & 3,822 & 4,922 & 197  \\ 
       \bottomrule	 
  \end{tabular}
        \caption{Characterization of our VizWiz-Captions dataset.  Shown is the average count per caption as well as the total count of unique words, nouns, verbs, adjectives, and spatial relation words for each dataset with respect to all captions, various subsets to support finer-grained analysis, and MSCOCO-Captions dataset~\cite{chen2015MicrosoftCOCOcaptions} for comparison. (adj = adjectives; spa-rel = spatial relations)}
        ~\label{table_captionAnalysis}
\end{table*} 

\paragraph{VizWiz-Captions Characterization.}  
Next, we characterized the caption content.  For this purpose, we excluded from our analysis all captions that contain the pre-canned text about insufficient quality images (``Quality issues are too severe...") as well as those that were rejected.  This resulted in a total of 168,826 captions. 

We first quantified the composition of captions, by examining the typical description length as well as the typical number of objects, descriptors, actions, and relationships.  To do so, we computed as a proxy the average number of words as well as the average number of nouns, adjectives, verbs, and spatial relation words per caption.  Results are shown in Table~\ref{table_captionAnalysis} (row 1).  Our findings reveal that sentences typically consist of roughly 13 words that involve four to five objects (i.e., nouns) in conjunction with one to two descriptors (i.e., adjectives), one action (i.e., verb), and two relationships (i.e., spatial relationship words).  Examples of sentences featuring similar compositions include ``A hand holding a can of Ravioli over a counter with a glass on it" and ``Red car parked next to a black colored SUV in an outside dirt parking lot."

We enriched our analysis by examining the typical caption composition separately for the 16\% (i.e., 6,339) of images that originated from a captioning use case and the remaining 84\% of images that originated from a VQA use case (meaning the image came paired with a question).  Results are shown in Table~\ref{table_captionAnalysis}, rows 2--3.  We observe that the composition of sentences is almost identical for both use cases.  This offers encouraging evidence that the images taken from a VQA setting are useful for large-scale captioning datasets.  

We further enriched our analysis by examining how the caption composition changes based on whether the image contains text.  We deemed an image as containing text if the majority of the five crowdworkers indicate it does.  In our dataset, 63\% (24,812) of the images contain text.  The caption compositions for both subsets are shown in Table~\ref{table_captionAnalysis}, rows 4--5.  Our findings reveal that images containing text tend to have more nouns and fewer adjectives than images that lack text.  Put differently, the presence of text appears to be more strongly correlated to the object recognition task.  We hypothesize this is in part because crowdworkers commonly employ both a generic object recognition category followed by a specific object category gleaned from reading the text when creating their descriptions; e.g., ``a box of Duracell procell batteries" and ``a can of Ravioli."  It's also possible that text is commonly present in more complex scenes that show a greater number of objects.

We also quantified the diversity of concepts in our dataset.  To do so, we report parallel analysis to that above, with a focus on the \emph{absolute number of unique} words, nouns, adjectives, verbs, and spatial relation words across all captions.  Results are shown in the right half of Table~\ref{table_captionAnalysis}.  These results demonstrate that the dataset captures a large diversity of concepts, with over 24,000 unique words.  We visualize the most popular words in the Supplementary Materials, and conduct further analysis below to offer insight into how these concepts relate/differ to those found in popular computer vision datasets.  

\paragraph{Comparison to Popular Captioning Dataset.}
We next compared our dataset to the popular MSCOCO-Captions dataset~\cite{chen2015MicrosoftCOCOcaptions}, and in particular the complete MSCOCO training set for which the captions are publicly-available.  

Paralleling our analysis of VizWiz-Captions, we quantified the average as well as total unique number of words, nouns, adjectives, verbs, and spatial words in MSCOCO-Captions~\cite{chen2015MicrosoftCOCOcaptions}.  To enable side-by-side comparison, we not only analyzed the entire MSCOCO-Captions training set but also randomly sampled the same number of images with the identical distribution of number of captions per image as was analyzed for VizWiz-Captions.  We call this subset MSCOCO-Sample. Results are shown in Table~\ref{table_captionAnalysis}, rows 6--7.  The results reveal that VizWiz-Captions tends to have a larger number of words per caption than MSCOCO-Captions; i.e., an average of 13 words versus 11.3 words.  This is true both for the full set as well as the sample from MSCOCO-Captions.  As shown in Table~\ref{table_captionAnalysis}, the greater number of words is due to a greater number of nouns, adjectives, and spatial relation words per caption in VizWiz-Captions.  Possible reasons for this include that the images show more complex scenes and that crowdworkers were motivated to provide more descriptive captions when knowing the target audience is people who are blind.  

We additionally measured the content overlap between the two datasets.  Specifically, we computed the percentage of words that appear both in the most common 3,000 words for VizWiz-Captions and the most common 3,000 words in  MSCOCO-Captions.  The overlap is 54.4\%.  This finding underscores a considerable domain shift in the content that blind photographers take pictures of and what artificially constructed datasets represent.  We visualize examples of novel concepts not found in MSCOCO-Captions in the Supplementary Materials.   

We also assessed the similarity of captions generated by different humans using the specificity score~\cite{jas2015Imagespecificity} for both our dataset and MSCOCO-Captions.  Due to space constraints, we show the resulting distributions of scores in the Supplementary Materials for both datasets.  In summary, the scores are similar.

\paragraph{Comparison to Visual Question Answering Dataset.}
Given that 84\% of the images originate from a VQA use case (i.e., where a question was also submitted about the image), our new dataset offers a valuable test bed to explore the potential for generic image captions to answer users' real visual questions.  Accordingly, we explore this for each image in our dataset for which we both have publicly-available answers for the question and the question is deemed to be ``answerable"~\cite{gurari2019VizWizPrivDatasetRecognizing,gurari2018VizWizGrandChallenge}. 

\begin{table}[t!]
  \centering
        \begin{tabular}{ l  c  c  c  c  c  c  c  c  c  c  c  c }
     \toprule
       & \multicolumn{4}{c}{\bf All Images} & \multicolumn{4}{c}{{\bf Images With Text}} & \multicolumn{4}{c}{{\bf Images Without Text}} \\
     \cmidrule(r){2-5} \cmidrule(r){6-9} \cmidrule(r){10-13}
     & All & Yes/No & \# & Other & All & Yes/No & \# & Other & All & Yes/No & \# & Other \\ 
    \midrule
      Quant & 33\% & 1\% & 8\% & 34\% & 35\% & 1\% & 10\% & 36\% & 30\% & 1\% & 3\% & 31\%\\ 
      Qual & 32\% & 35\% & 23\% & 38\% & -- & -- & -- & -- & -- & -- & -- & --\\ 
       \bottomrule	 
  \end{tabular}
        \caption{Percentage of VQAs for which an image caption contains the answer with respect to both a quantitative (``Quant") and qualitative (``Qual") analysis.  Fine-grained quantitative analysis is shown based on the type of answer that is elicited by the visual question (i.e., ``yes/no", ``\#", and ``other") as well as based on whether the images contain text. (\# = number)}
        ~\label{table_vqaAnalysis}
\end{table} 

We first evaluate this using a \emph{quantitative measure}.  Specifically, for the 24,842 answerable visual questions in the publicly-available training and validation splits, we tally the percentage for which the answer can be found in at least one of the five captions using exact string matching.  We set the answer to the most popular answer from the 10 provided with each visual question.  We conduct this analysis with respect to all images as well as separately for only those images which are paired with different answer types for the visual questions---i.e., ``yes/no" (860 images), ``number" (314 images), and ``other" (23,668 images).  Results are shown in Table~\ref{table_vqaAnalysis}.  Overall, we observe that captions contain the information that people who are blind were seeking for roughly one third of their visual questions.  This sets a lower bound, since string matching is an extremely rigid scheme for determining whether text matches.  

We perform parallel quantitative analysis based on whether images contain text.  For visual questions that contain text (i.e., 15,910 answerable visual questions), we again analyze the visual questions that lead to ``yes/no" (447 images), ``number" (218 images), and ``other" (15,245 images) answers.  We also perform this analysis on only the subset of visual questions that lack text (i.e., 8,932 answerable visual questions)---i.e., ``yes/no" (413 images), ``number" (96 images), and ``other" (8,423 images).  Results are shown in Table~\ref{table_vqaAnalysis}.  We observe that the answer tends to be contained in the caption more often when the image contains text.  This discrepancy is the largest for ``number" questions, which we hypothesize is due to images showing currency.  People seem to naturally want to characterize how much money is shown for such images, which conveniently is the information sought by those asking the questions.  


 
 
To also capture when the answer to a visual question is provided implicitly in the captions, we next used a \emph{qualitative approach}.  We sampled 300 visual questions, with 100 for each of the three answer types.\footnote{For ``yes/no" visual questions, we sampled 50 that have the answer ``yes" and another 50 with the answer ``no."  For ``number" visual questions, we sampled 50 that begin with the question ``How many" and another 50 that begin with ``How much."  Finally, we randomly sampled another 100 visual questions from the ``other" category.}  Then, one of the authors reviewed each visual question with the answers and five captions to decide whether each visual question was answered by any of the captions about the image.  Results are shown in Table~\ref{table_vqaAnalysis}, row 4.  We observe a big jump in percentage for ``yes/no" and ``number" questions.  The greatest boost is observed for ``yes/no" visual questions where the percentage jumps from 0\% to 35\%.  We attribute this to the ``yes" questions more than the ``no" questions--i.e., 22/50 for ``yes" and 13/50 for ``no"---since content that is asked about may be described when it is present in the image but will almost definitely not be described when it is not.  Still, ``no" questions often arise because, when the answer can be inferred, the caption typically also answers a valuable follow-up question.  For example, a caption that states ``A carton of banana flavored milk sits in a clear container with eggs" arguably answers the question ``Is this chocolate milk?" (i.e., the answer is ``no") while providing additional information (i.e., it is ``banana milk").  

Altogether, our findings show that at least one third of the visual questions can be answered with image captions.  In other words, the captions regularly provide useful information for people who are blind.  We attribute this large percentage partly to the fact that many questions for VQA just paraphrase a request to complete the image captioning task; e.g., nearly half of the questions ask a variant of ``what is this" or ``describe this"~\cite{gurari2018VizWizGrandChallenge}.  It also may often be obvious to the people providing captions what information the photographer was seeking when submitting the image with a question.  Regardless of the reason though, it appears the extra work of devising a question regularly can be unnecessary in practice.  A valuable direction for future research is to continue improving our understanding for how to align captions with real end users' interests.

\paragraph{Comparison to Popular Image Classification Datasets.}
Observing that automated captioning algorithms often build off of pretrained modules that perform more basic tasks such as image classification and object detection (e.g., trend dates back at least to Baby Talk~\cite{kulkarni2013BabytalkUnderstandinggenerating} in 2013), we next examine the overlap between concepts in VizWiz-Captions and popular vision datasets that often are used to train such modules.  For our analysis, we focus on three visual tasks: recognizing objects, scenes, and attributes.  

We began by tallying how many popular concepts from existing vision datasets for the three vision tasks are found in VizWiz-Captions.  To do so, we computed matches using extract string matching.  When comparing concepts in VizWiz-Captions to the object categories that span both ImageNet~\cite{russakovsky2015Imagenetlargescale} and COCO~\cite{lin2014MicrosoftcocoCommon}, we found that all nine categories that are shared across the two datasets are also found in VizWiz-Captions.  Similarly, we found that all scene categories which span both SUN~\cite{xiao2010SundatabaseLargescale} and Places205~\cite{zhou2014Learningdeepfeatures} (i.e., 70 categories) are captured in VizWiz-Captions.  Additionally, all attribute categories that span both COCO-Attributes~\cite{patterson2016CocoattributesAttributes} and SUN-Attributes~\cite{patterson2012Sunattributedatabase} (i.e., 14 categories) are captured in VizWiz-Captions.  Consequently, across all three tasks, all concepts that are shared across the pair of mainstream vision datasets are also present in VizWiz-Captions.  This is interesting in part because VizWiz-Captions was not created with any of these tasks in mind.  Moreover, it underscores the promise for models trained on existing datasets to generalize well in recognizing some content encountered by people who are blind in their daily lives.

\begin{figure*}[t!]
\centering
    \begin{subfigure}{0.32\textwidth}
    \includegraphics[width=\linewidth]{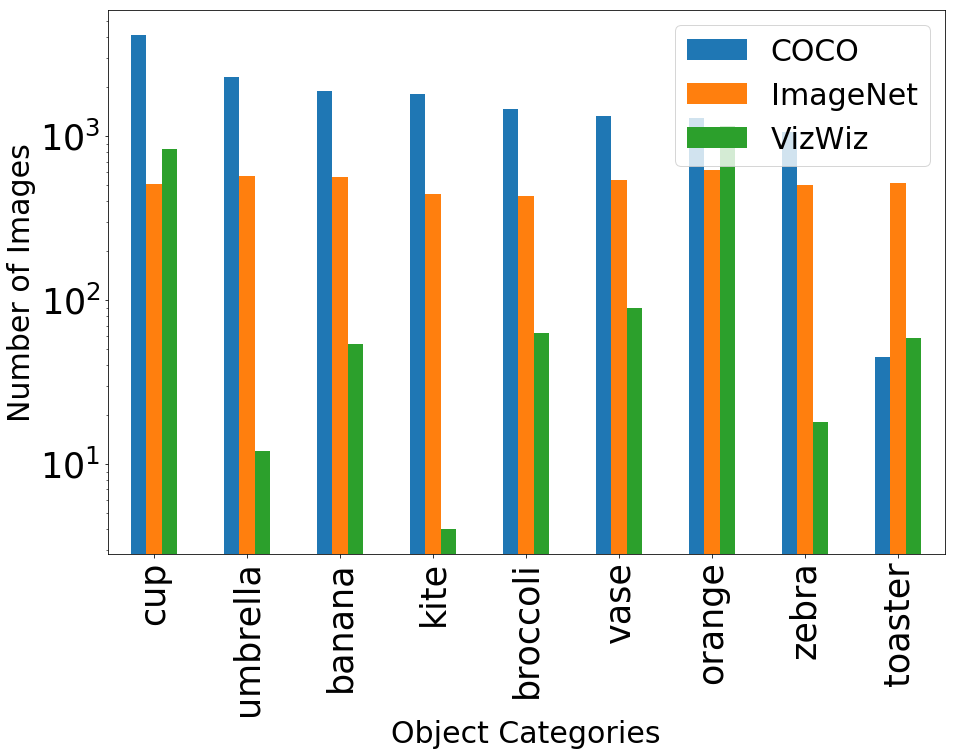}
    \caption{} \label{fig:2a}
    \end{subfigure}
    \hfill 
	\begin{subfigure}{0.32\textwidth}
    \includegraphics[width=\linewidth]{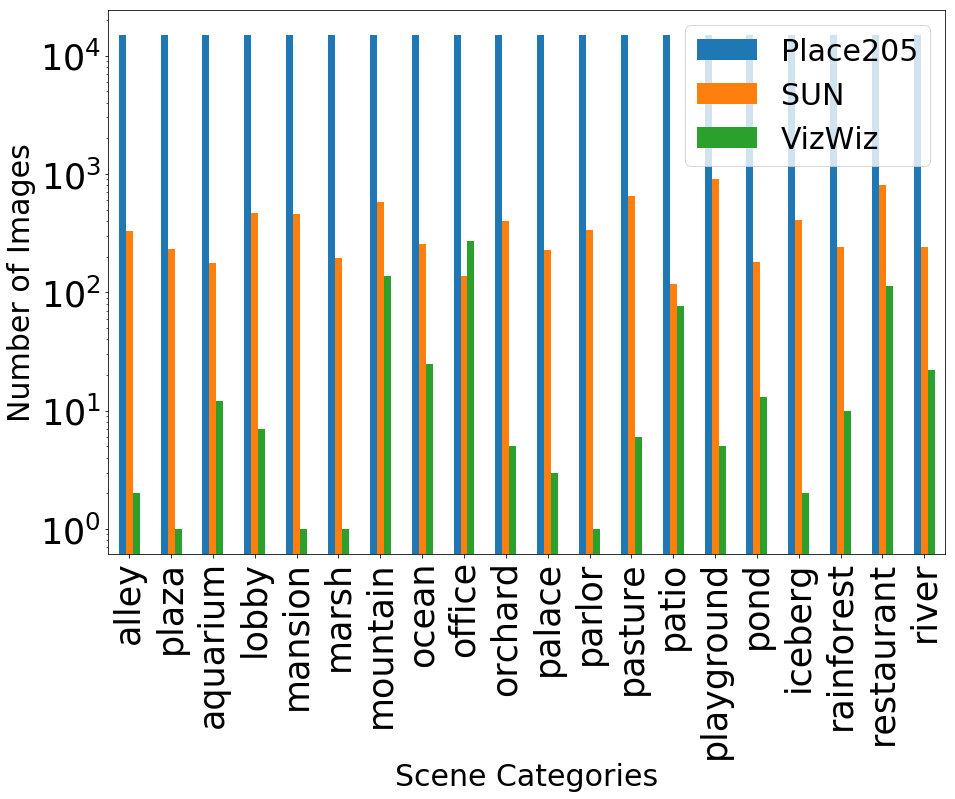}
    \caption{} \label{fig:2b}
    \end{subfigure}
    \hfill 
    \begin{subfigure}{0.32\textwidth}
    \includegraphics[width=\linewidth]{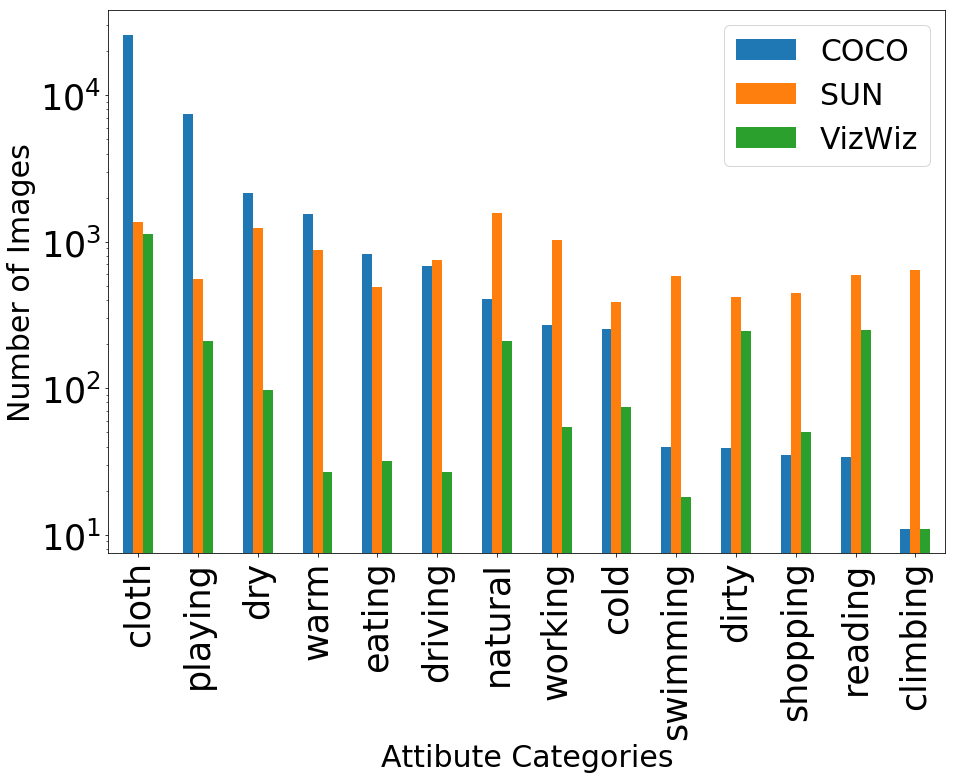}
    \caption{} \label{fig:2c}
    \end{subfigure}
    \hfill
  \caption {Histogram showing how many images in our VizWiz-Captions as well as popular vision datasets contain each category for the following vision problems: (a) object recognition, (b) scene classification, and (c) attribute recognition.}
  \label{fig_imageClassificationComparison}
\end{figure*}

We also tally how many of the images in each dataset contain the popular concepts discussed above.  Results are reported with respect to each of three classification tasks in Figure~\ref{fig_imageClassificationComparison}.\footnote{We show parallel analysis in the Supplementary Materials using the proportions of each dataset rather than absolute numbers.  For both sets of results, we only show a subset of the 70 scene categories.}  As shown, the number of examples in VizWiz-Captions is typically considerably fewer than observed for the other two popular datasets per task.  This is not entirely surprising given that the absolute number of images in VizWiz-Captions is at least an order of magnitude smaller than most of the datasets (i.e., the object and scene classification datasets).  We offer this analysis as a lower bound since \emph{explicitly} asking crowdworkers whether each category is present could reveal a greater prevalence of these concepts.  Observing that relying on data from real use cases alone likely provides an insufficient number of examples per category to successfully train algorithms, this finding highlights a potential benefit of contrived datasets in supplementing examples to our real-world dataset.  We leave this idea as a valuable area for future work.

We also computed the percentage of all categories from each of the classification datasets that are captured by VizWiz-Captions.  Again, we used exact string matching to do so.  For object recognition, VizWiz-Captions contains only 1\% of the categories in ImageNet and 11\% of those in COCO.  For scene recognition, VizWiz-Captions contains only 18\% of the categories in SUN and 34\% of those in Places205.  For attribute recognition, VizWiz-Captions contains only 14\% of the categories in COCO-Attributes and 7\% of those in SUN-Attributes.  Observing that these vision datasets are reserved to a range of hundreds to at most a thousand categories while we know from Table~\ref{table_captionAnalysis} that VizWiz-Captions contains thousands of unique nouns and adjectives, these datasets appear to provide very little coverage for the diversity of content captured in VizWiz-Captions.  Altogether, these findings offer promising evidence that existing contrived image classification datasets provide a considerable mismatch to the concepts encountered by blind users who are trying to learn about their visual surroundings.  Our findings serve as an important reminder that much progress is still needed to accommodate the diversity of content found in real-world settings.

%% file: algorithm-benchmarking.tex
\section{Algorithm Benchmarking}
\label{sec_algBenchmarking}
We next benchmarked state-of-art image captioning algorithms to gauge the difficulty of VizWiz-Captions and what makes it difficult for modern algorithms.

\paragraph{Dataset Splits.}
Using the same test set as prior work~\cite{gurari2018VizWizGrandChallenge}, we applied roughly a 70\%/10\%/20\% split for the train/val/test sets respectively, resulting in a 23,431/7,750/8,000 split.  To focus algorithms on learning novel captions, we exclude from training and evaluation captions with pre-canned text about insufficient quality images or rejected ones that were deemed spam.

\paragraph{Baselines.}
We benchmarked nine algorithms based on three modern image captioning algorithms that have been state-of-art methods for the MSCOCO-Captions~\cite{chen2015MicrosoftCOCOcaptions} challenge: Up-Down~\cite{Anderson2017up-down}, SGAE~\cite{yang2019auto}, and AoANet~\cite{huang2019attention}.  Up-Down~\cite{Anderson2017up-down} combines bottom-up and top-down attention mechanisms to consider attention at the level of objects and other salient image regions.  SGAE~\cite{yang2019auto} relies on a Scene Graph Auto-Encoder (SGAE) to incorporate language bias into an encoder-decoder framework, towards generating more human-like captions.  AoANet~\cite{huang2019attention} employs an Attention on Attention (AoA) module to determine the relevance between attention results and queries.  We evaluated all three algorithms, which originally were trained on the MSCOCO-Captions dataset, \emph{as is}.  These results are useful in assessing the effectiveness of the MSCOCO training dataset for teaching computers to describe images taken by people who are blind.  We also \emph{fine-tuned} each pretrained network to VizWiz-Captions and \emph{trained each network from scratch} on VizWiz-Captions.  These algorithms are helpful for assessing the usefulness of each model architecture for describing images taken by people who are blind.  For all algorithms, we used the publicly-shared code and default training hyper-parameters reported by the authors.  We also benchmarked a commercial text detector\footnote{\url{https://docs.microsoft.com/en-us/azure/cognitive-services/computer-vision/concept-recognizing-text}} on test images containing text.   

\paragraph{Evaluation.}
We evaluated each method with eight metrics that often are used for captioning:  BLEU-1-4~\cite{papineni2002BLEUmethodautomatic}, METEOR~\cite{denkowski:lavie:meteor-wmt:2014}, ROUGE-L~\cite{lin2004rouge}, CIDEr-D~\cite{vedantam2015CiderConsensusbasedimage}, and SPICE~\cite{anderson2016SpiceSemanticpropositional}.

\begin{table*}[t!]
  \centering
        \begin{tabular}{ c  c  c  c  c  c  c  c  c  c}
    
    \toprule
       && \bf B@1 & \bf B@2 & \bf B@3 & \bf B@4 & \bf METEOR & \bf ROUGE & \bf CIDEr & \bf SPICE  \\
     \midrule
     \multirow{3}{*}{{{\bf \cite{Anderson2017up-down}~}}} 
        & pretrained & 52.8 & 32.8 & 19.2 & 11.3 & 12.6 & 35.8 & 18.9 & 5.8 \\ 
        & from scratch & 64.1 & 44.6 & 30.0 & 19.8 & 18.4 & 43.2 & 49.7 & 12.2 \\
        & fine-tuned & 62.1 & 42.3 & 28.2 & 18.6 & 18.0 & 42.0 & 48.2 & 11.6 \\
    \hline
    \multirow{3}{*}{{{\bf \cite{yang2019auto}~}}} 
        & pretrained & 55.8 & 36.0 & 21.8 & 13.5 & 13.4 & 38.1 & 20.2 & 5.9 \\ 
        & from scratch & 67.3 & 48.1 & 33.2 & 22.8 & 19.4 & 46.6 & 52.4 & 12.8 \\
        & fine-tuned & \textbf{68.5} & \textbf{49.4} & \textbf{34.5} & \textbf{23.9} & 20.2 & \textbf{47.3} & \textbf{61.2} & 13.5 \\
    \hline
    \multirow{3}{*}{{{\bf \cite{huang2019attention}~}}} 
        & pretrained & 54.9 & 34.7 & 21.0 & 13.2 & 13.4 & 37.6 & 19.4 & 6.2 \\ 
        & from scratch & 66.4 & 47.9 & 33.4 & 23.2 & \textbf{20.3} & 47.1 & 60.5 & \textbf{14.0} \\
        & fine-tuned & 66.6 & 47.4 & 32.9 & 22.8 & 19.9 & 46.6 & 57.6 & 13.7 \\
    \bottomrule
    \end{tabular}
        \caption{Performance of top-performing image captioning algorithms on the VizWiz-Captions test set with respect to eight metrics.  Results are shown for three variants of the algorithms: when they are pre-trained on MSCOCO-Captions~\cite{chen2015MicrosoftCOCOcaptions}, trained only on the VizWiz-Captions dataset, and pre-trained on MSCOCO-Captions followed by fine-tuning to the VizWiz-Captions dataset. (B@ = BLEU-)}
        ~\label{table_algBenchmarking}
\end{table*}

\begin{figure*}[t!]
\centering
\includegraphics[width=1.0\textwidth]{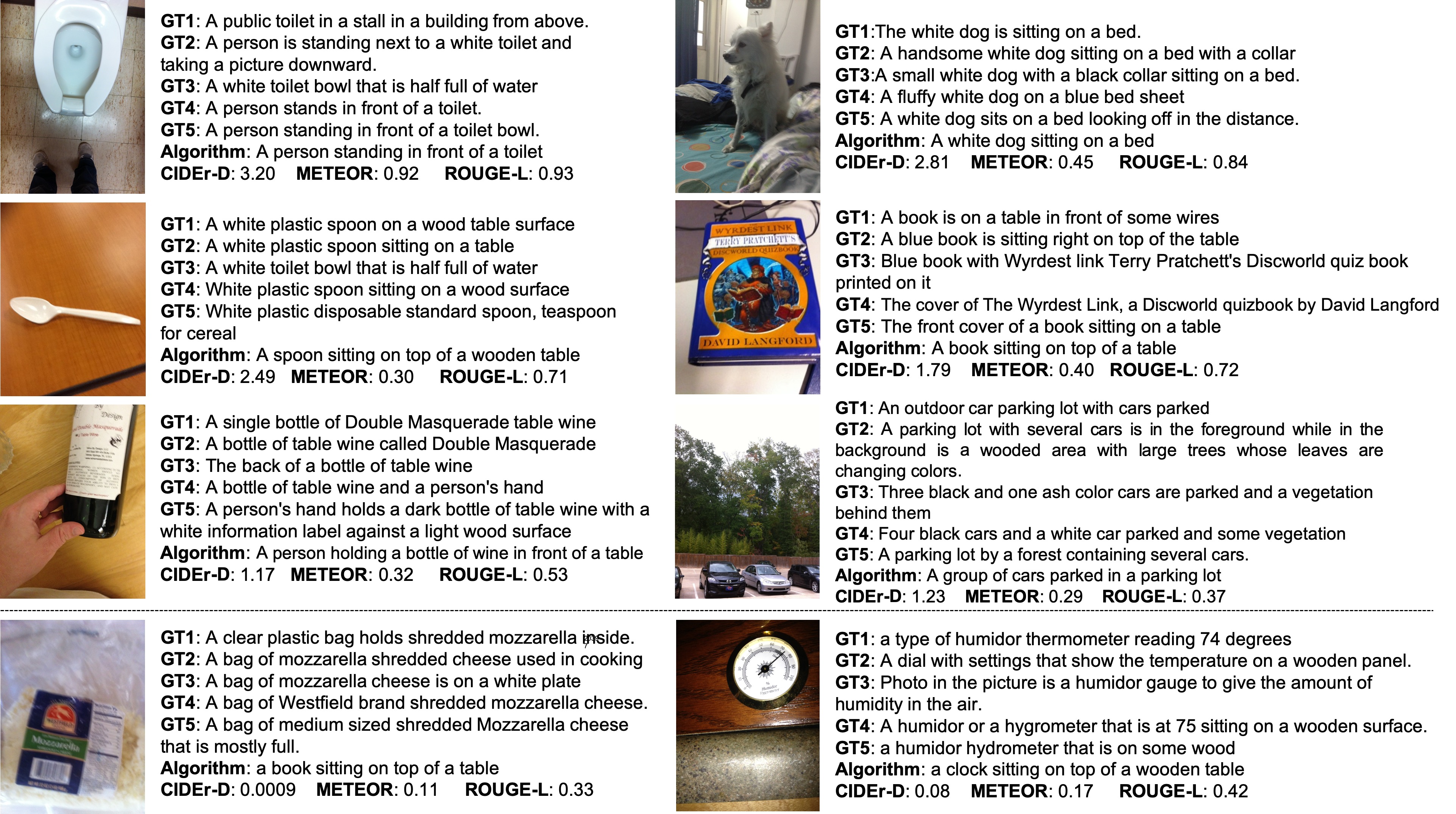}
\caption{Examples of a state-of-art image captioning algorithm's successes (top three rows) and failures (bottom row) in generating captions for images taken in a real use case.  Results are for SGAE~\cite{yang2019auto} pretrained on MSCOCO-Captions.}
\label{fig_pretrained}
\end{figure*}

\paragraph{Overall Performance.}
We report the performance of each method in Table~\ref{table_algBenchmarking}.

Observing the performance of existing algorithms that are \emph{pretrained} on MSCOCO-Captions~\cite{chen2015MicrosoftCOCOcaptions}, we see that they can occasionally accurately predict captions for images coming from blind photographers.  This is exciting as it shows that progress on artificially-constructed datasets can translate to successes in real use cases.  We attribute the prediction successes to when the images are both good quality and show objects that are common in MSCOCO-Captions, as exemplified in the top six examples in Figure~\ref{fig_pretrained}. 

We consistently observe considerable performance improvements from the algorithms when training them on VizWiz-Captions, including when they are trained from scratch and fine-tuned.  For instance, we observe roughly a 10 percentage point boost with respect to BLEU-1 and 30 percentage point boost with respect to CIDEr-D across the three algorithms.  Still, the scores are considerably lower than what is observed when these same algorithmic frameworks are evaluated on the MSCOCO-Captions test set.  For example, we observe the BLEU-1 score is over 20 percentage points lower and the METEOR score is almost 20 percentage points lower, when comparing the performance of the top-performing algorithm for VizWiz-Captions against the top-performing algorithm for MSCOCO-Captions (i.e., AoANet~\cite{huang2019attention}).  This finding highlights that VizWiz-Captions currently offers a challenging dataset for the vision community.  

When comparing outcomes between algorithms that are trained from scratch on VizWiz-Captions versus fine-tuned to VizWiz-Captions, we do not observe a considerable difference.  For instance, we observe better performance when Up-Down~\cite{Anderson2017up-down} and AoANet~\cite{huang2019attention} are trained from scratch on VizWiz-Captions rather than fine-tuned from models pretrained on MSCOCO-Captions, and vice versa for SGAE~\cite{yang2019auto}.  We found it surprising there is similar performance, given that VizWiz-Captions is roughly one order of magnitude smaller than MSCOCO-Captions.  Valuable areas for future work include investigating the benefit of domain adaptation methods as well as how to successfully leverage larger contrived datasets (e.g., MSCOCO-Captions) to improve the performance of algorithms on VizWiz-Captions.  

\paragraph{Fine-Grained Analysis.}
We enriched our analysis to better understand why algorithms struggle to accurately caption the images.  To do so, we evaluated the top-performing algorithms for VizWiz-Captions with respect to two characteristics.  First, we characterized performance independently for images in the test set based on whether they are flagged as containing text.  We also characterized performance independently for images flagged as different difficulty levels, based on the number of crowdworkers who deemed the images insufficient quality to generate a meaningful caption; i.e., easy is when all five people generated novel captions, medium when 1-2 crowdworkers flagged the images as insufficient quality for captioning, and difficult when 3-4 crowdworkers flagged the images as insufficient quality.  Results are shown in Table~\ref{table_fineGrainedAlgBenchmarking} for the top algorithms from Table~\ref{table_algBenchmarking}; i.e., ``from scratch" for ~\cite{Anderson2017up-down} and ~\cite{huang2019attention} and ``fine-tuned" for ~\cite{yang2019auto}.

\begin{table*}[t!]
  \centering
        \begin{tabular}{ c  c  c  c  c  c  c  c  c  c}
    
    \toprule
       && \bf B@1 & \bf B@2 & \bf B@3 & \bf B@4 & \bf METEOR & \bf ROUGE & \bf CIDEr & \bf SPICE  \\
     \midrule
     \multirow{5}{*}{{{\bf \cite{Anderson2017up-down}~}}}  
        & WithText & 65.7 & 46.6 & 32.3 & 21.7 & 19.2 & 45.1 & 49.3 & 12.6 \\ 
        & LackText & 60.2 & 40.0 & 25.2 & 15.7 & 16.9 & 39.7 & 46.7 & 11.4 \\
        \cdashline{2-10}
        & Easy & 67.8 & 48.3 & 33.2 & 22.2 & 19.5 & 45.7 & 53.2 & 12.5 \\
        & Medium & 60.8 & 40.0 & 25.4 & 16.2 & 17.2 & 40.4 & 45.8 & 11.9 \\
        & Difficult & 32.1 & 16.9 & 9.2 & 5.4 & 10.6 & 26.2 & 34.7 & 9.0 \\
    \hline
    \multirow{5}{*}{{{\bf \cite{yang2019auto}~}}} 
        & WithText & 69.9 & 51.2 & 36.6 & 25.8 & 21.0 & 49.3 & 62.2 & 14.1 \\ 
        & LackText & 65.7 & 45.9 & 30.3 & 20.2 & 18.7 & 43.7 & 55.5 & 12.5 \\
        \cdashline{2-10}
        & Easy & 72.2 & 53.3 & 38.0 & 26.7 & 21.4 & 50.0 & 65.8 & 13.9 \\
        & Medium & 65.1 & 44.7 & 29.3 & 19.3 & 18.8 & 44.3 & 55.8 & 13.3 \\
        & Difficult & 35.6 & 20.1 & 11.7 & 7.4 & 11.9 & 28.8 & 42.7 & 9.3 \\
    \hline
    \multirow{5}{*}{{{\bf \cite{huang2019attention}~}}} 
        & WithText & 68.3 & 50.1 & 35.8 & 25.3 & 21.3 & 49.2 & 62.7 & 14.6 \\ 
        & LackText & 63.0 & 43.7 & 28.8 & 18.9 & 18.5 & 43.3 & 52.5 & 12.9 \\
        \cdashline{2-10}
        & Easy & 69.8 & 51.4 & 36.5 & 25.6 & 21.4 & 49.6 & 64.3 & 14.3 \\
        & Medium & 63.6 & 44.0 & 29.2 & 19.5 & 19.2 & 44.5 & 56.0 & 14.2 \\
        & Difficult & 36.0 & 20.3 & 11.8 & 7.6 & 12.2 & 29.6 & 44.3 & 10.6 \\
    \hline
    Text\_API & WithText & 14.9 & 8.8 & 5.7 & 3.9 & 10.4 & 15.9 & 24.6 & -- \\
    \bottomrule
  \end{tabular}
        \caption{Analysis of the top-performing image captioning algorithms and a text detection algorithm based on whether images contain text and the image ``difficulty". (B@ = BLEU-)}
        ~\label{table_fineGrainedAlgBenchmarking}
\end{table*} 

We observe two trends for the performance of algorithms based on whether text is present.  We find the text detector does very poor, underscoring a key challenge for designing algorithms is to figure out how to integrate knowledge about text into captions.  In contrast, we find that all captioning algorithms perform better when text is present.  Initially, we found this surprising given that none of the benchmarked algorithms were designed to handle text (e.g., by incorporating an optical recognition module). Moreover, images with text cover many more unique concepts than images lacking text, as shown in Table~\ref{table_captionAnalysis}. We hypothesize the improved performance is because images containing text provide a simpler domain that conforms to a fewer set of templates for the captions.  For example, from visual inspection, we observe captions for such images often include ``a box/bag of ... on/in ...".  The captioning patterns for this simpler domain may be easier to learn for the algorithms.  If so, this underscores an inadequacy of current evaluation metrics and a need for new metrics that prioritize the information people who are blind want.  


When observing algorithm performance based on the captioning difficulty level, we find it parallels human difficulty with algorithms performing best on the easiest images for humans. While not surprising, this finding underscores the practical difficulty of designing algorithms that can handle low quality images, which we know are somewhat common from real users of image captioning services (i.e., people who are blind).  

%% file: conclusions.tex
\section{Conclusions}
We offer VizWiz-Captions as a valuable foundation for designing image captioning algorithms to support a natural, socially-important use case.  More broadly, our analysis reveals important problems that the vision community needs to address in order to deliver more generalized algorithms.  Interesting future work includes holistically improving vision solutions to include consideration of potentially, valuable additional sensors to more effectively meet real users' needs (e.g., GPS, sound waves, infrared). 


%% file: supplementary-materials.tex
\appendix
\noindent {\LARGE \textbf{Appendix}}
\vspace{1em}

This document supplements Sections 3 and 4 of the main paper.  In particular, it includes the following:

\begin{itemize}
\item Implementation description of the crowdsourcing system (supplements \textbf{Section 3.1})
\item Analysis of the consistency of captions collected from different crowd workers for each image (supplements \textbf{Section 3.1})
\item Examples of images that are deemed insufficient or low quality for captioning (supplements \textbf{Section 3.2}) 
\item Visualizations and quantitative analysis demonstrating the diversity of content in VizWiz-Captions and how it compares to that in MSCOCO-Captions and the image classification datasets (supplements \textbf{Section 3.2}) 
\item Algorithm performance when using data augmentation during training by blurring images (supplements \textbf{Section 4})  
\end{itemize}

\input{supp-dataset-creation}
\input{supp-dataset-analysis}
\input{supp-algorithm-benchmarking}

%% file: supp-dataset-creation.tex
\begin{figure*}[t!]
\centering
\includegraphics[width=0.9\textwidth]{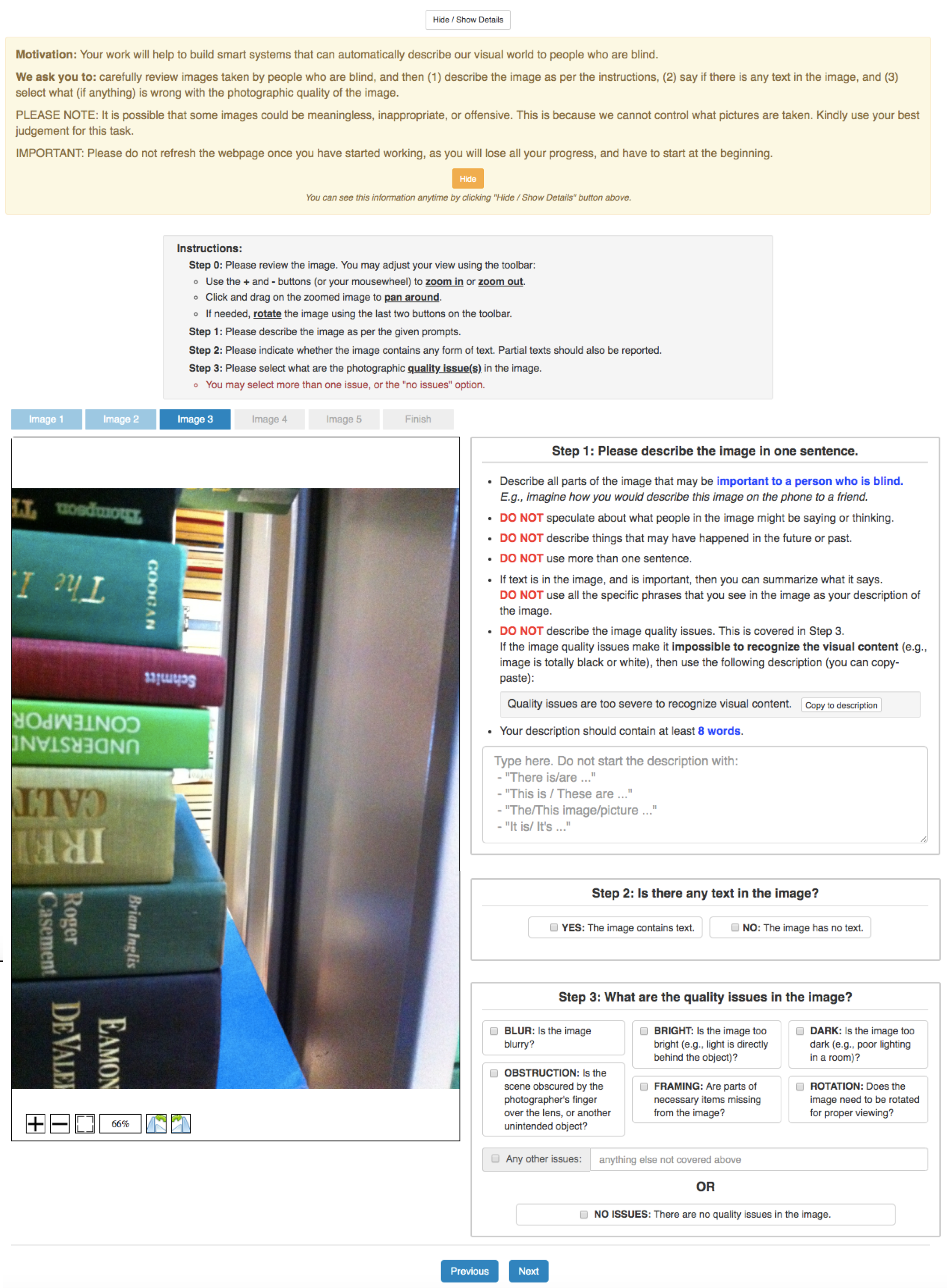}
\caption{Interface used to crowdsource the collection of image captions.}
\label{fig_taskInterface}
\end{figure*}

\section{Dataset Creation (supplements Section 3.1)}
\subsection{Crowdsourcing Task Design}
A screen shot of our crowdsourcing interface is shown in Figure~\ref{fig_taskInterface}.  The interface prevented the crowdworker from proceeding to the next image (for the sequential set of five images) or submitting the work until the following criteria was met for each image description:

\begin{itemize}
    \item Contains at least eight words (to encourage rich content)
    \item Contains only one period followed by a space (to restrict crowd worker to one-sentence descriptions)
    \item Sentences do not begin with the following prefixes (to discourage uninformative content): ``There is", ``There are", ``This is", ``These are", ``The image", ``The picture", ``This image", ``This picture", ``It is", and ``It's"
\end{itemize}

We collected all the annotations over five batches of Human Intelligence Tasks (HITs) in order to minimize the impact of inadequate workers.  After each batch, we identified workers who we considered to be problematic and blocked them from participating in subsequent batches.  To identify problematic workers, the authors reviewed a subset of the crowdworkers' results.  The following mechanisms were used to determine which workers' captions to review:

\begin{itemize}
    \item Workers who were a statistical outlier in time-to-submit (taking either too little or too much time) by 1.95 times the standard deviation for all the results
    \item Workers who used CAPS LOCK for more than 50\% of the caption text
    \item Workers who used the canned text ("Quality issues are too severe to recognize visual content") for more than 50\% of the images that they captioned
    \item Workers who were the only one to either use or not use the canned text ("Quality issues are too severe to recognize visual content") for an image
    \item Workers who used words like "quality", "blur" and "blurry" (but not the canned text), and so were not focusing on content in the image
    \item Random sample from all results
\end{itemize}

We also included numerous additional quality control mechanisms.  First, crowdworkers could not submit their results until their work passed an automated check that verified they followed a number of the task instructions, including writing at least 8 words, providing only one sentence, and not starting the description with ``There is..." or other unsubstantial starting phrases.  We also only accepted crowdworkers who previously had completed over 500 HITs with at least a 95\% acceptance rate.  We will publicly-share the crowdsourcing code to support reproducibility of this interface.

\subsection{Caption Post-processing}
We employed Microsoft Azure's spell-checking API\footnote{https://azure.microsoft.com/en-us/services/cognitive-services/spell-check/} to find and correct misspelled words in the submitted captions.  We chose this approach because we found from initial testing that it outperforms other tested methods, including because it can recognize brand names (which are common in our dataset).  It also does a good job of correcting grammar and capitalizing words when appropriate (e.g. changing "dell" to "Dell").  When spell-checking all captions which are neither canned text nor from blocked workers (i.e., 169,073 captions), 14\% (i.e, 23,424) were flagged as containing unknown "tokens" (aka - words).  We replace each unknown ``token" with the most confidently recommended word suggested by the Azure API.

%% file: supp-dataset-analysis.tex
\begin{figure*}[b!]
\centering
    \begin{subfigure}{0.49\textwidth}
    \includegraphics[width=\textwidth]{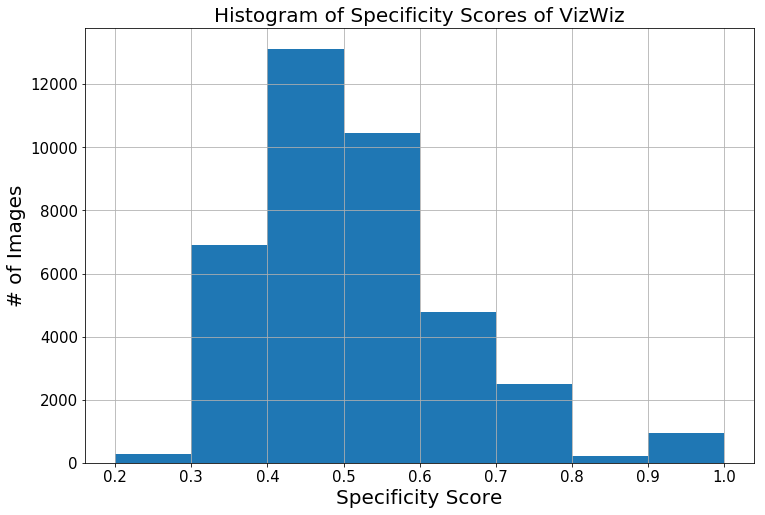}
    \caption{VizWiz-Captions}
    \end{subfigure}
    \hfill 
	\begin{subfigure}{0.49\textwidth}
    \includegraphics[width=\textwidth]{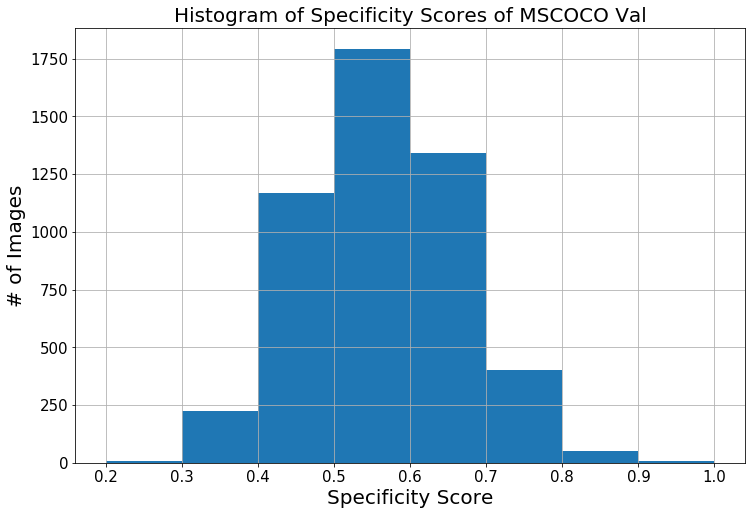}
    \caption{MSCOCO-Captions}
    \end{subfigure}
    \hfill 
\caption{Distribution of image specificity scores for images in (a) VizWiz-Captions and (b) MSCOCO-Captions.}
\label{fig_specificityScores}
\end{figure*}

\section{Caption Consistency (supplements Section 3.2)}
We report the distributions of specificity scores that indicate the similarity between the five captions per image generated by different humans for all images in our VizWiz-Captions dataset and the MSCOCO-Captions validation set independently.  Scores range between 0 and 1, with numbers closer to 1 indicating greater consistency between the five captions per image.  Results are shown in Figure~\ref{fig_specificityScores}.  

\begin{figure}[b!]
\centering
\includegraphics[width=\textwidth]{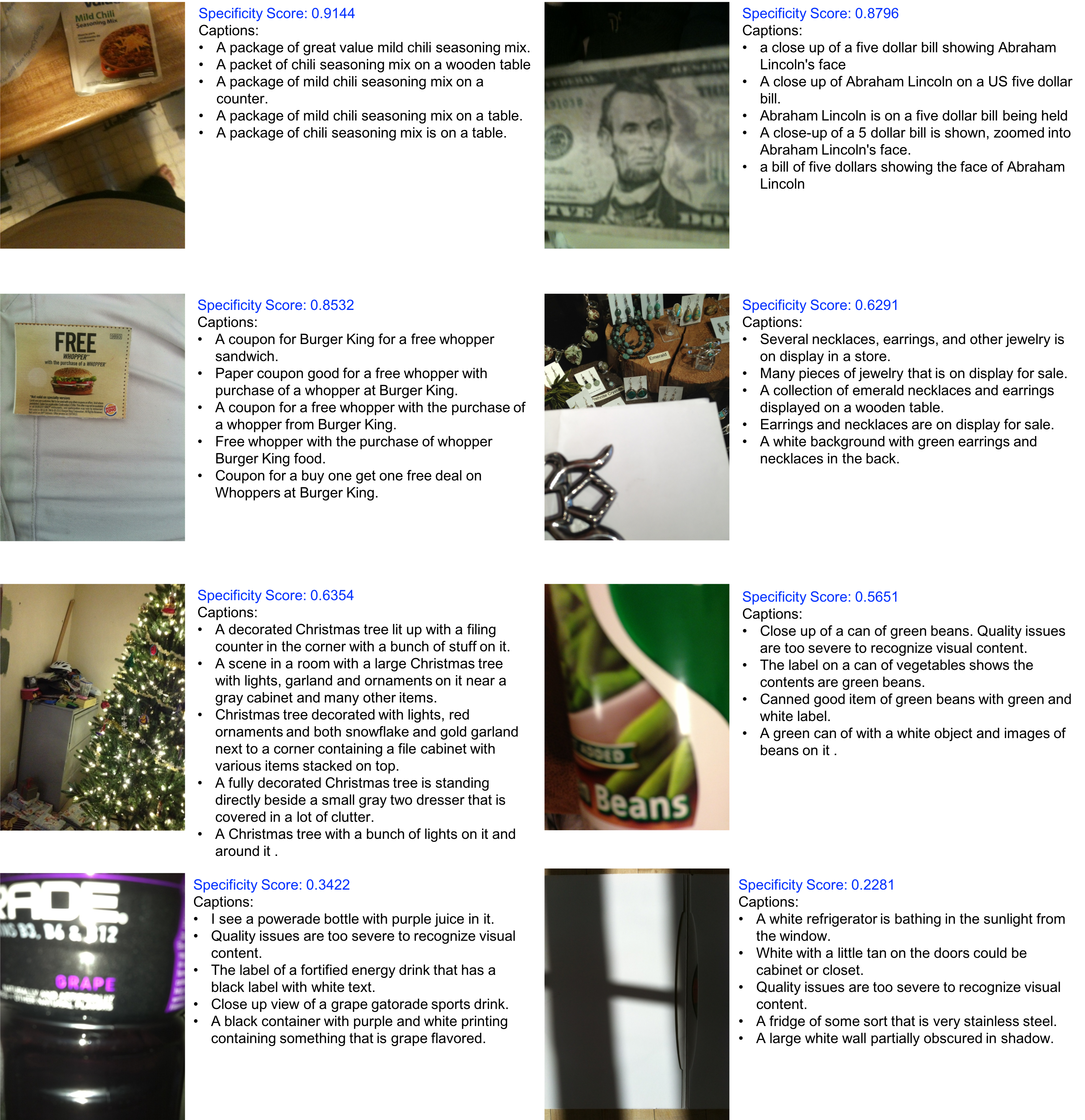}
\caption{Examples of images that lead to a range of specificity scores when analyzing the captions collected from different crowdworkers.}
\label{fig_qualSpecificityScores}
\end{figure}

While the distributions are similar overall, we observe scores are skewed more towards 0 for VizWiz-Captions.  We attribute these slightly greater annotation differences to annotators providing different level of detail and different types of detail, as exemplified in Figure 3 in the main paper. We show examples of the diversity of captions that arise from different crowdworkers for a range of specificity scores in Figure~\ref{fig_qualSpecificityScores}.

\section{Dataset Analysis (supplements Section 3.2)}
\label{sec_datasetAnalysis}

\subsection{Insufficient Quality Images}
\begin{figure}[h!]
\centering
\includegraphics[width=0.8\textwidth]{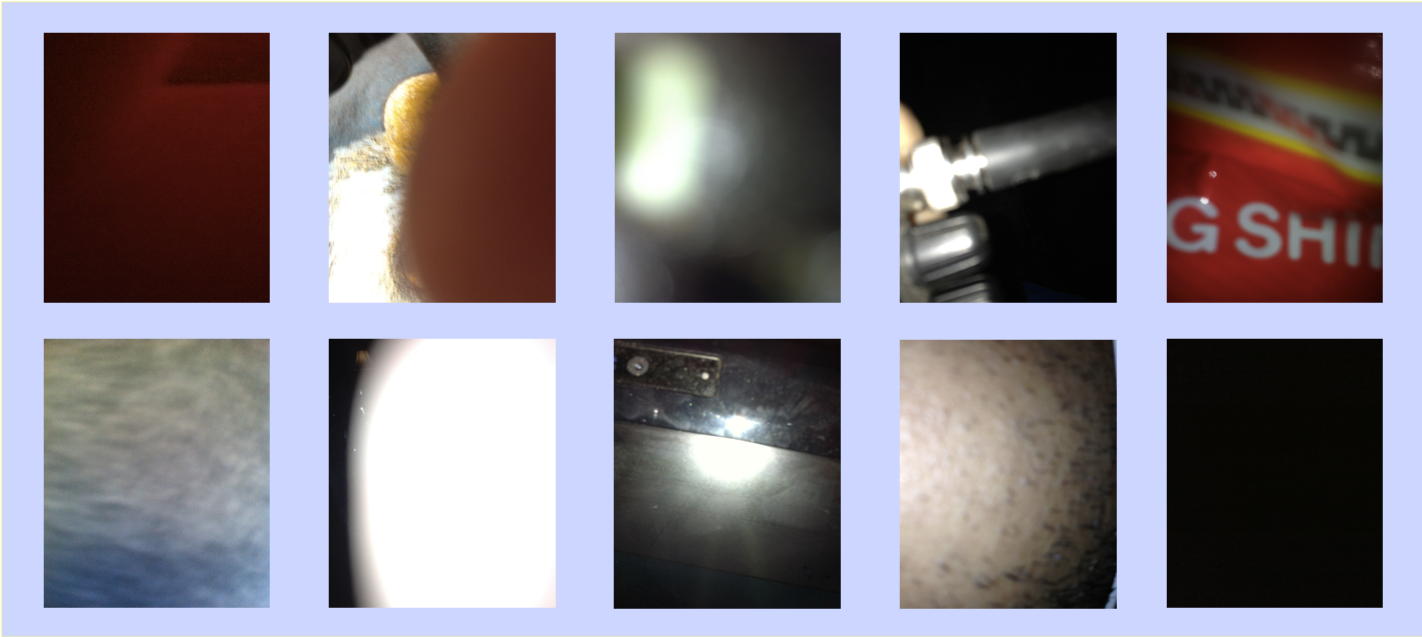}
\caption{Examples of images that are unanimously labeled as insufficient quality to be captioned.}
\label{fig_insufficientQualityExamples}
\end{figure}

Figure~\ref{fig_insufficientQualityExamples} exemplifies images that were deemed insufficient or lower quality for captioning based on the agreement of five crowdworkers.

\begin{figure*}[b!]
\centering
    \begin{subfigure}{0.49\textwidth}
    \includegraphics[width=\textwidth]{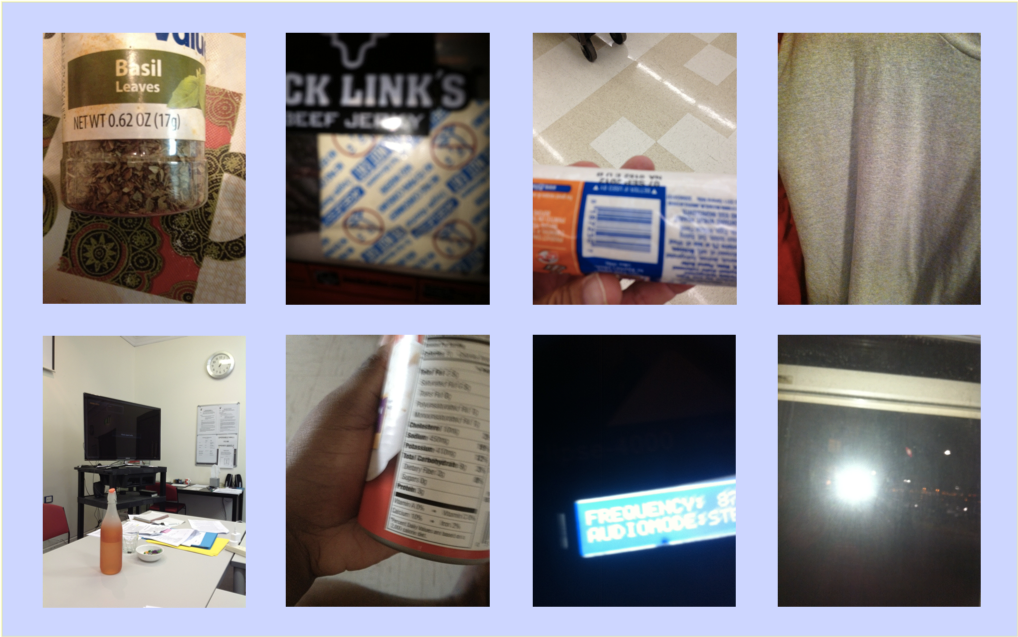}
    \caption{Medium difficulty for image captioning}
    \end{subfigure}
    \hfill 
	\begin{subfigure}{0.49\textwidth}
    \includegraphics[width=\textwidth]{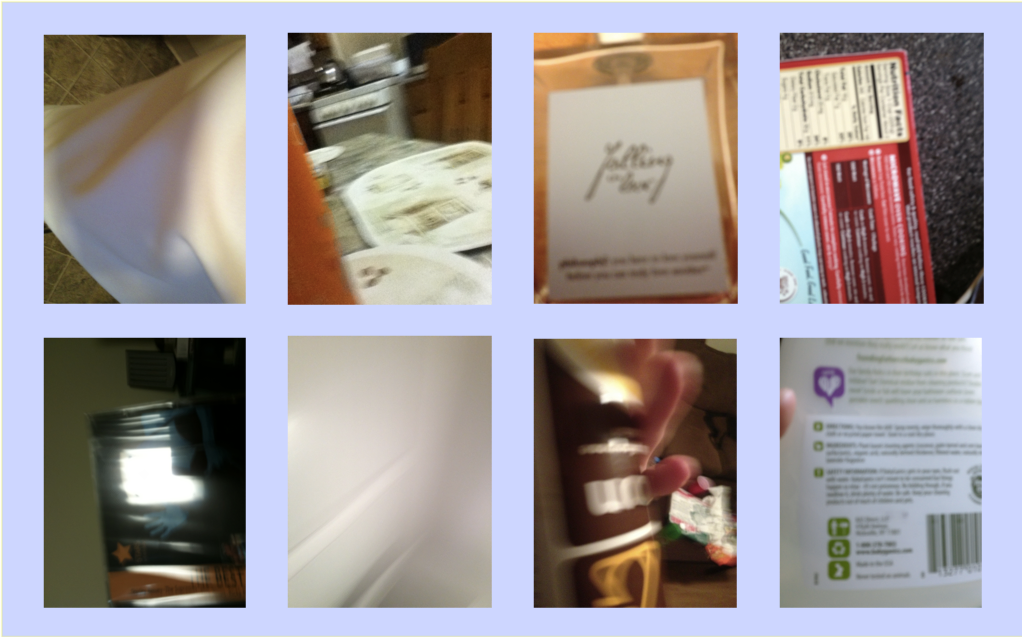}
    \caption{High difficulty for image captioning}
    \end{subfigure}
    \hfill 
\caption{Examples of images that are labeled as insufficient quality by (a) 1-2 crowdworkers and (b) 3-4 crowdworkers.}
\label{fig_lowQualityExamples}
\end{figure*}

We show examples of images that we deem medium or high difficulty based on the agreement of five crowdworkers who indicated the image is insufficient quality to generate a meaningful caption (i.e., 1-2 for medium and 3-4 for high difficulty) in Figure~\ref{fig_lowQualityExamples}.

\clearpage
\begin{figure*}[t!]
\centering
    \begin{subfigure}{0.32\textwidth}
    \includegraphics[width=\textwidth]{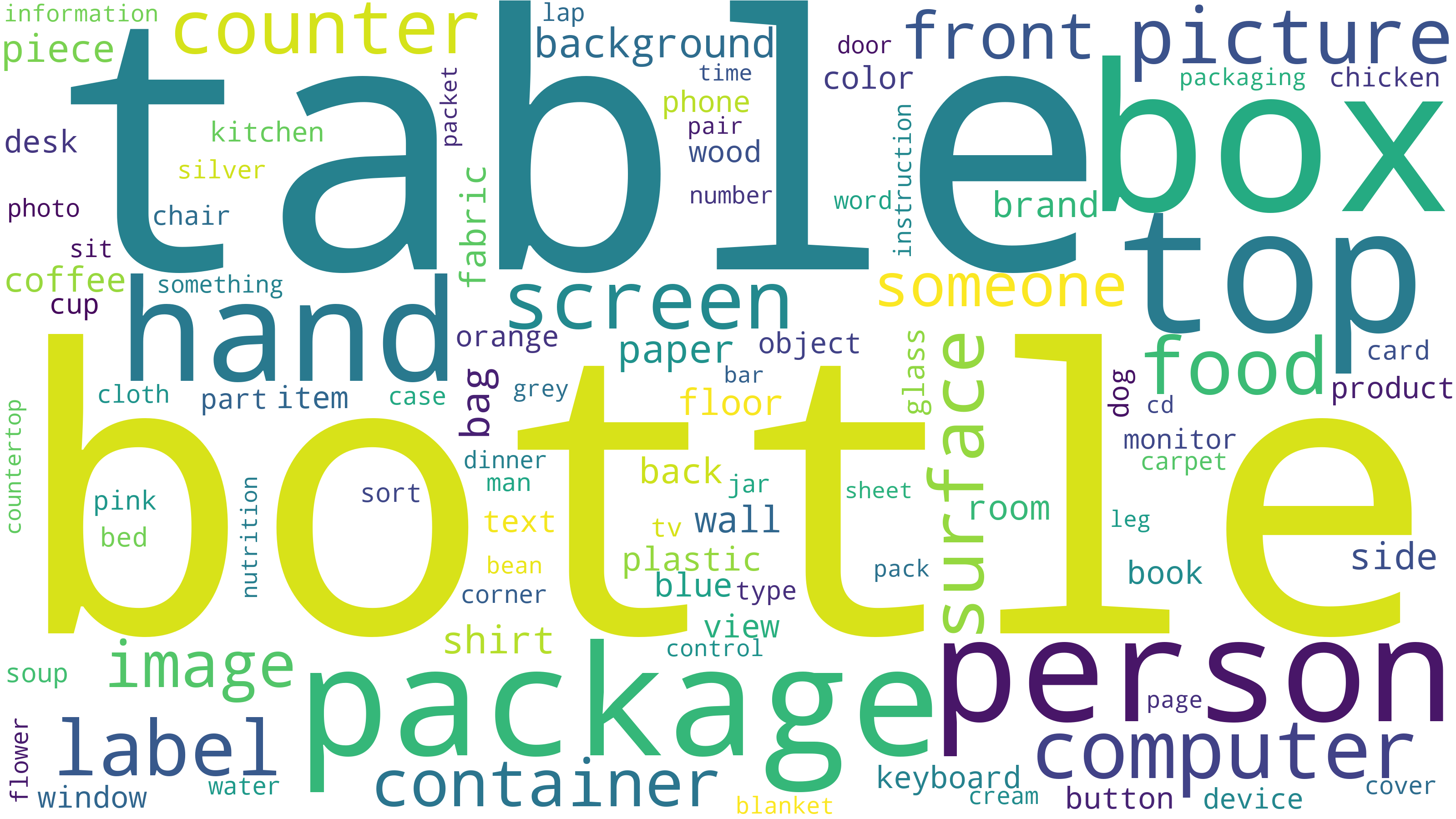}
    \caption{VizWiz-Captions}
    \end{subfigure}
    \hfill 
	\begin{subfigure}{0.32\textwidth}
    \includegraphics[width=\textwidth]{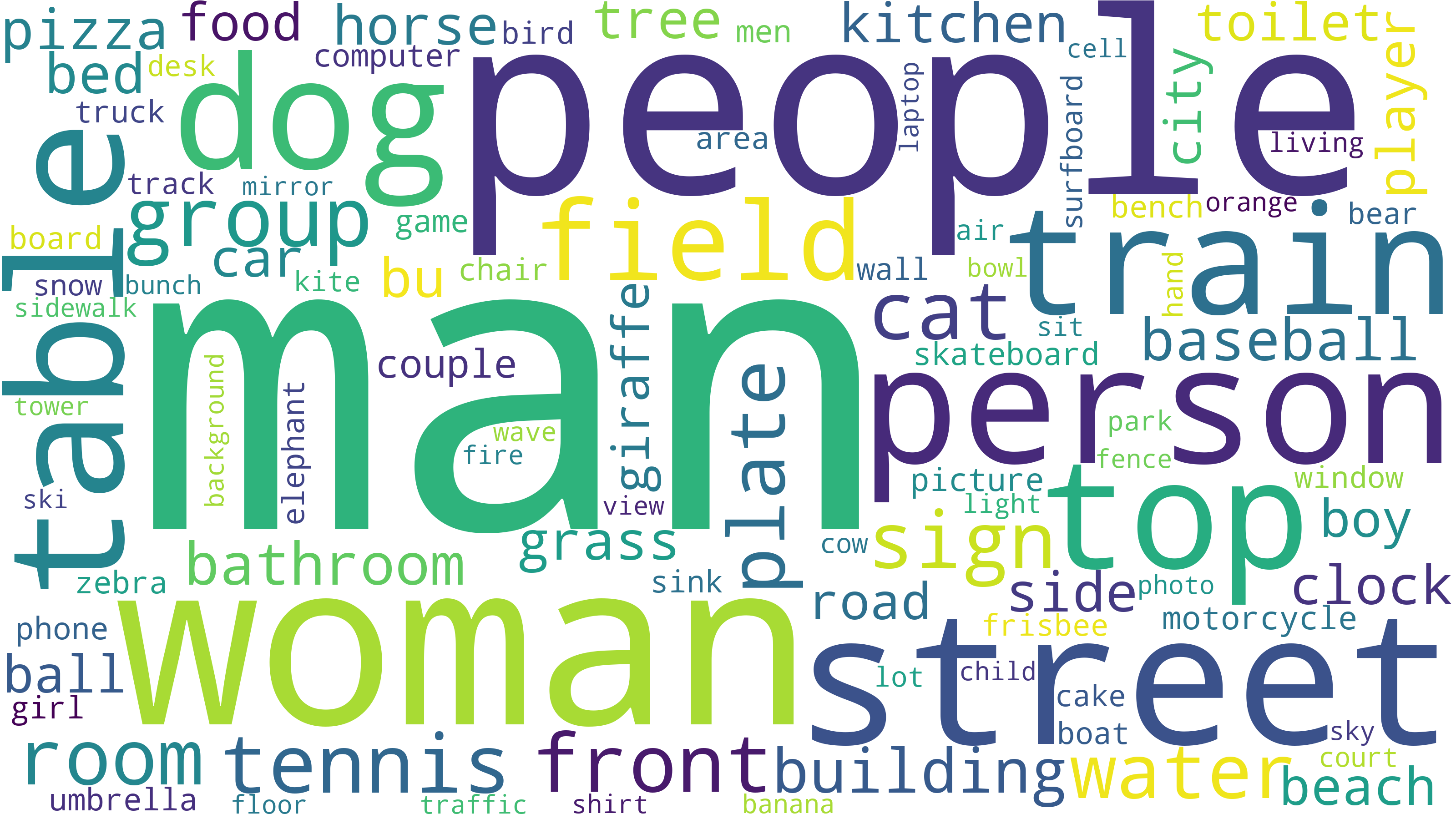}
    \caption{MSCOCO-Captions}
    \end{subfigure}
    \hfill 
    \begin{subfigure}{0.32\textwidth}
    \includegraphics[width=\textwidth]{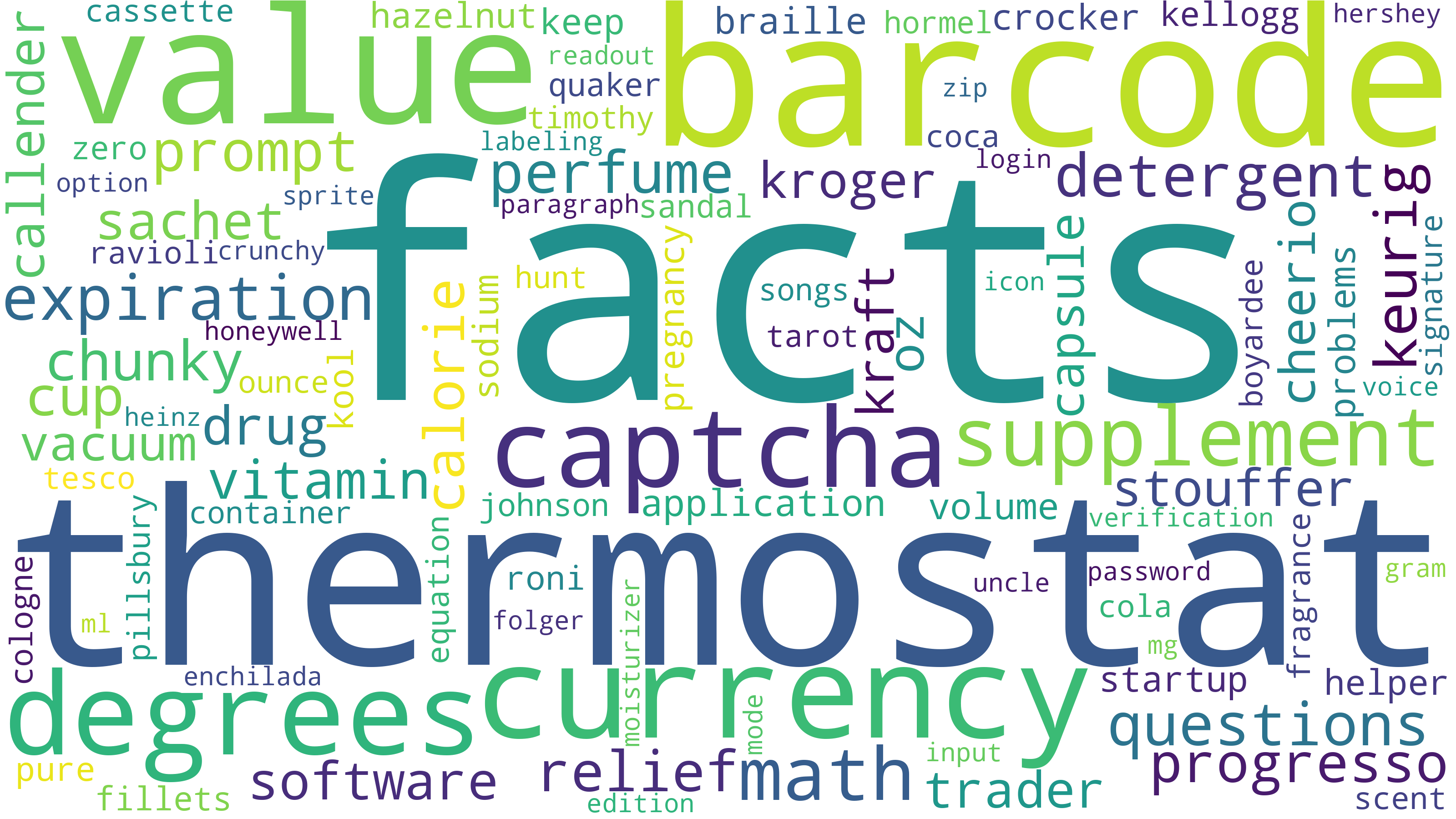}
    \caption{VizWiz-Captions only}
    \end{subfigure}
    \hfill 
\caption{Wordclouds for the most popular 100 nouns across all captions that are in (a) VizWiz-Captions, (b) MSCOCO-Captions, and (c) in VizWiz-Captions but not in MSCOCO-Captions.}
\label{fig_topNouns}
\end{figure*}

\begin{figure*}[t!]
\centering
    \begin{subfigure}{0.32\textwidth}
    \includegraphics[width=\textwidth]{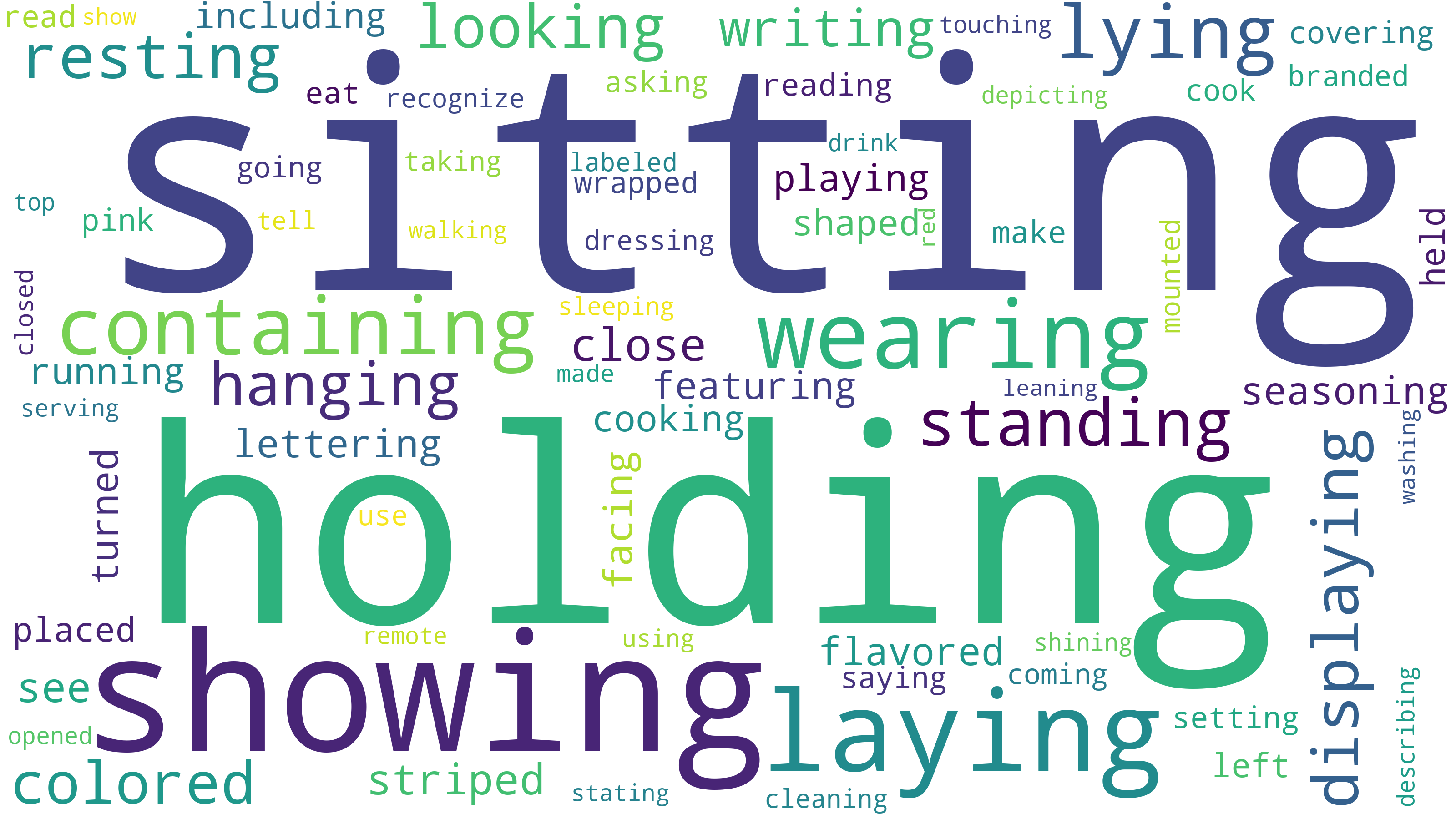}
    \caption{VizWiz-Captions}
    \end{subfigure}
    \hfill 
	\begin{subfigure}{0.32\textwidth}
    \includegraphics[width=\textwidth]{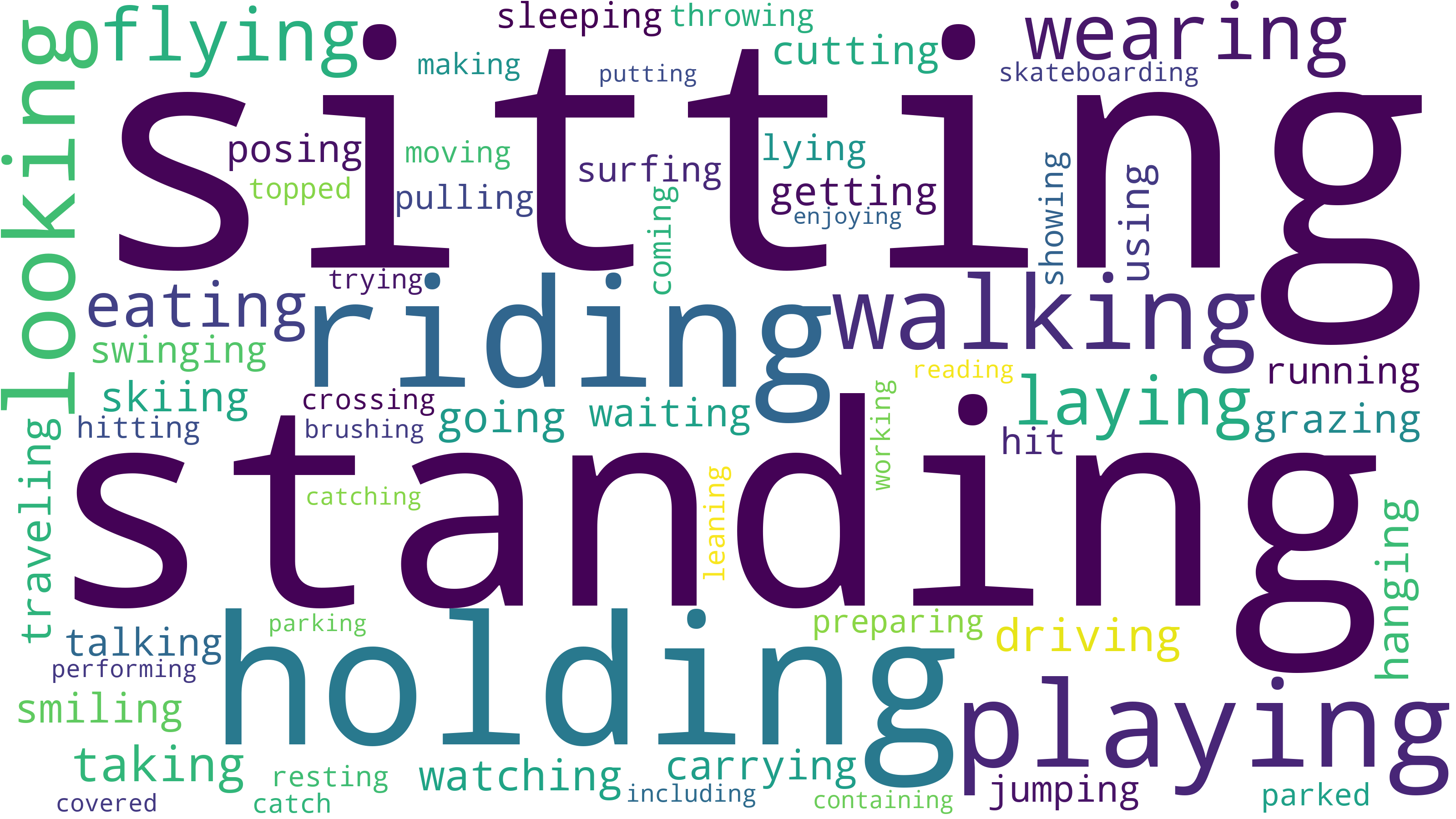}
    \caption{MSCOCO-Captions}
    \end{subfigure}
    \hfill 
    \begin{subfigure}{0.32\textwidth}
    \includegraphics[width=\textwidth]{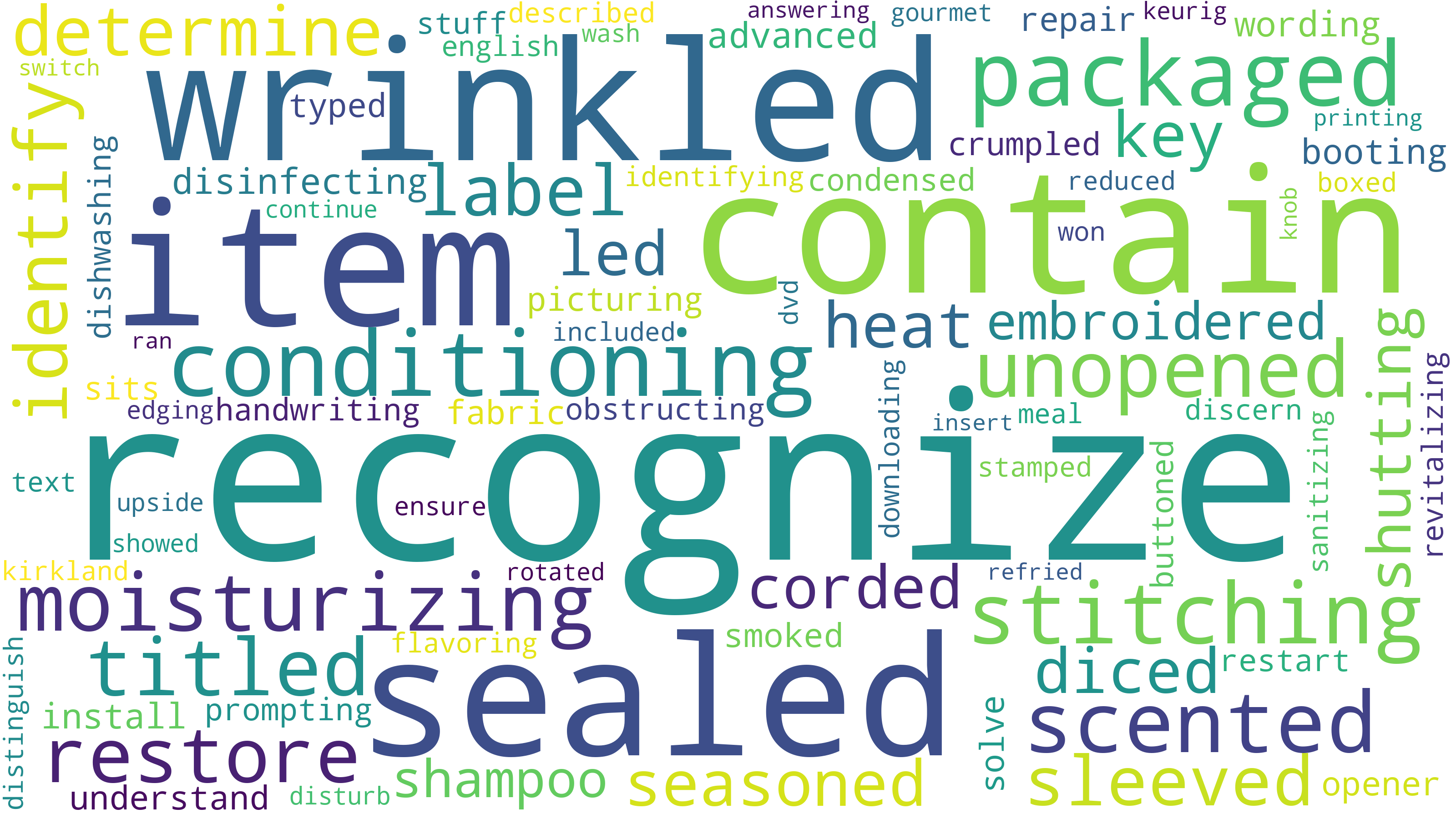}
    \caption{VizWiz-Captions only}
    \end{subfigure}
    \hfill 
\caption{Wordclouds for the most popular 100 verbs across all captions that are in (a) VizWiz-Captions, (b) MSCOCO-Captions, and (c) in VizWiz-Captions but not in MSCOCO-Captions.}
\label{fig_topVerbs}
\end{figure*}

\begin{figure*}[t!]
\centering
    \begin{subfigure}{0.32\textwidth}
    \includegraphics[width=\textwidth]{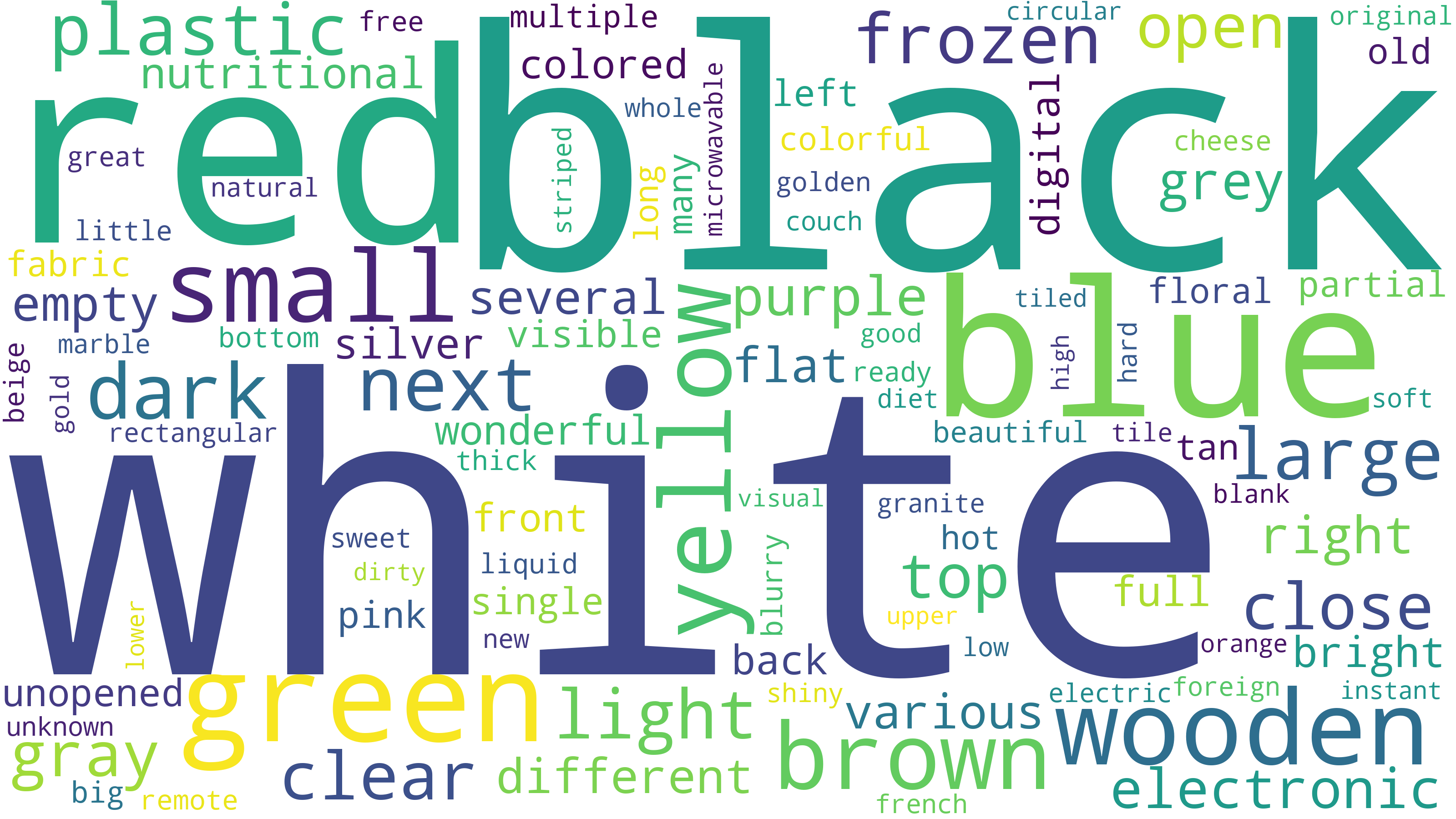}
    \caption{VizWiz-Captions}
    \end{subfigure}
    \hfill 
	\begin{subfigure}{0.32\textwidth}
    \includegraphics[width=\textwidth]{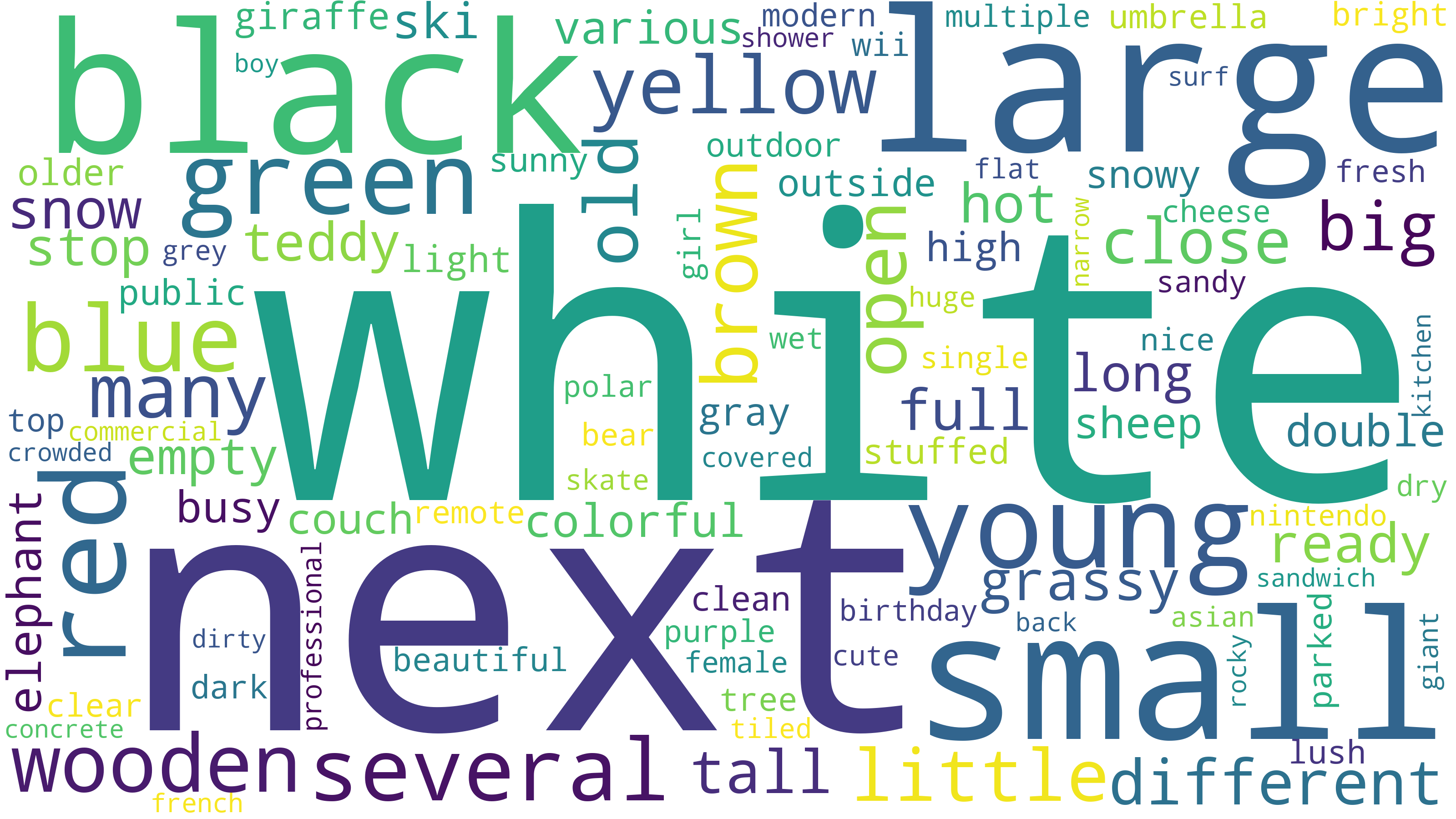}
    \caption{MSCOCO-Captions}
    \end{subfigure}
    \hfill 
    \begin{subfigure}{0.32\textwidth}
    \includegraphics[width=\textwidth]{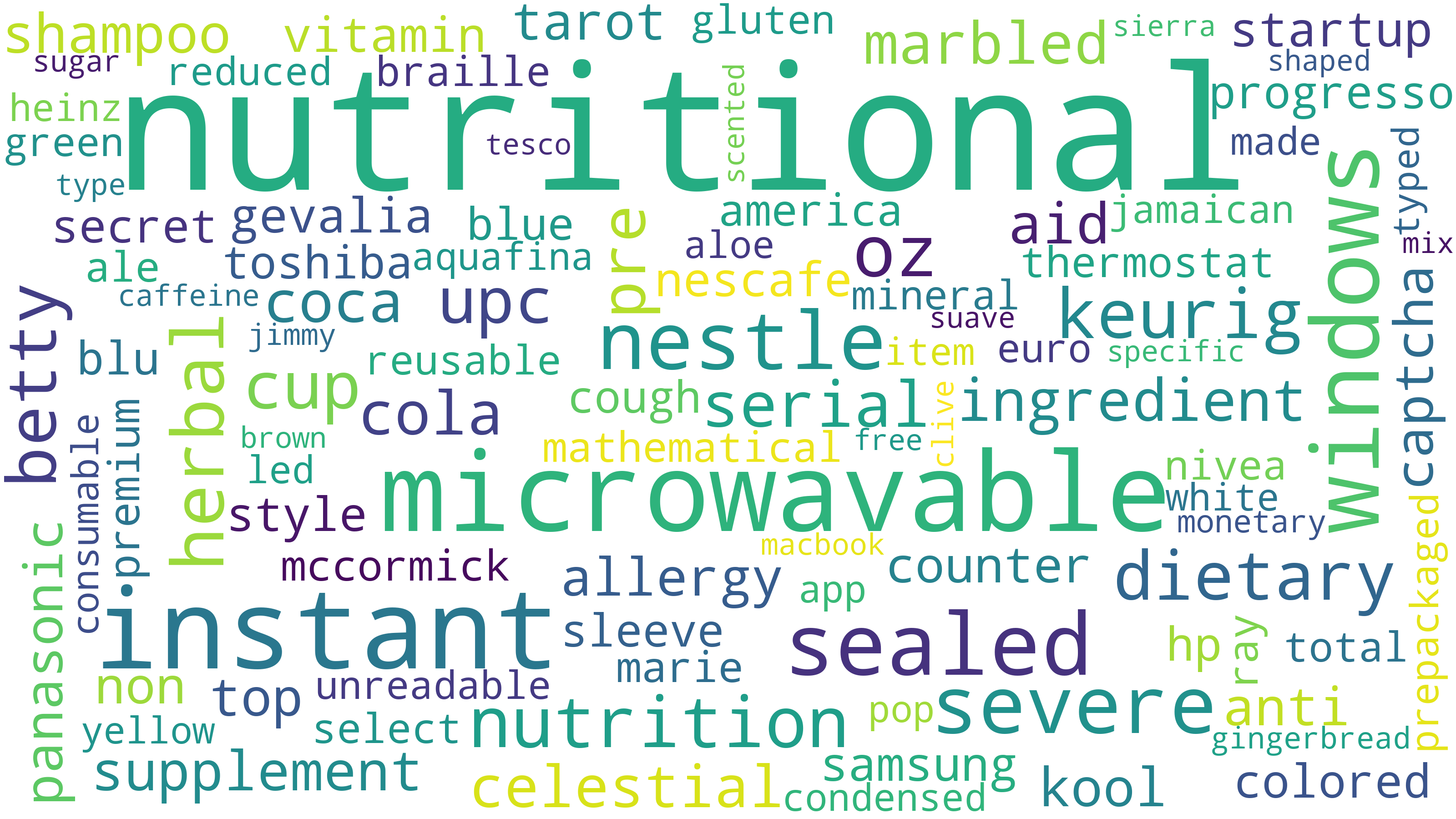}
    \caption{VizWiz-Captions only}
    \end{subfigure}
    \hfill 
\caption{Wordclouds for the most popular 100 adjectives across all captions that are in (a) VizWiz-Captions, (b) MSCOCO-Captions, and (c) in VizWiz-Captions but not in MSCOCO-Captions.}
\label{fig_topAdjs}
\end{figure*}

\subsection{Caption Characterization}
We visualize the most popular words included in the captions for each of the following word types analyzed in Table 1 of the main paper: nouns, verbs, and adjectives.  We do so both for VizWiz-Captions and MSCOCO-Captions to support comparison to today's mainstream captioning dataset.  For each word type we show the most common 100 words in VizWiz-Captions and MSCOCO-Captions separately as well as the most common 100 words that are in VizWiz-Captions but not found in MSCOCO-Captions.  Results for nouns, verbs, and adjectives are shown in Figures~\ref{fig_topNouns}, \ref{fig_topVerbs}, and \ref{fig_topAdjs} respectively.  We observe in Figure~\ref{fig_topNouns} that many popular words focus on items from daily living such as `table', `person', 'box', `food', and `monitor'.  Nouns that are absent from MSCOCO-Captions and popular in VizWiz-Captions largely focus on text and numbers (Figure~\ref{fig_topNouns}c), including `expiration', `captcha', `thermostat', `password', and `currency'. 

We also quantify the extent to which people are present in the images, given the high prevalence in many mainstream vision datasets. When tallying how many images are captioned using words related to people (i.e., people, person, man, woman, child, hand, foot, torso), the result is 27.6\% (i.e., 10,805/39,181).  When applying a person detector~\cite{ren2015faster}\footnote{We employed a faster-rcnn model pretrained on COCO. It has a ResNeXt-101 as the backbone and Feature Pyramid Network (FPN) to deal with objects of different scales. We only used the ``person" category out of the 80 categories. We filtered the detections with a threshold of 0.3, meaning we only counted a detection as valid if the confidence score is above 0.3.}, people are detected for 29.4\% (i.e., 11,499/39,181) of images.  We suspect this latter result is slightly larger than observed for captions because of person detections on background objects, such as newspaper or TV screens.  We suspect crowdworkers found such person detections to be insufficiently salient to be described as part of the captions.  Altogether, the relatively low prevalence of humans may in part be attributed to the fact that any images showing people's faces were filtered from the publicly-shared dataset to preserve privacy.  We suspect that the presence of people in our VizWiz-Captions compared to popular vision datasets will differ in that either (1) only parts of people appear in our images, such as only hands, legs, and torsos or (2) when the full body is shown, often it is because the person is on the cover of media (book, magazine, cd, dvd) or a product box.



We next report the percentage of overlap between the most common 3,000 words in VizWiz-Captions and the most common 3,000 words in MSCOCO-Captions for all words as well as with respect to each of the following word types: nounds, adjectives, and verbs.  Results are shown in Table~\ref{table_contentMatchBetweenDatasets}.  The higher percentage across all words than for the different word types is likely because there are many common stopwords that are shared across both datasets that do not belong to each word type.  

\begin{table}[h!]
\centering
\begin{tabular}{ l l l l  }
\toprule
\cmidrule(r){1-4}
words & nouns & adj & verbs  \\ 
\midrule
54.4\% & 45.1\% & 31.6\% & 42.8\%  \\ 
\bottomrule
\end{tabular}
\caption{Percentage of overlap between most common 3,000 words in VizWiz-Captions and the most common 3,000 words in MSCOCO-Captions.}
~\label{table_contentMatchBetweenDatasets}
\vspace{-0.5em}
\end{table}

\begin{figure*}[h!]
\centering
    \begin{subfigure}{0.32\textwidth}
    \includegraphics[width=\linewidth]{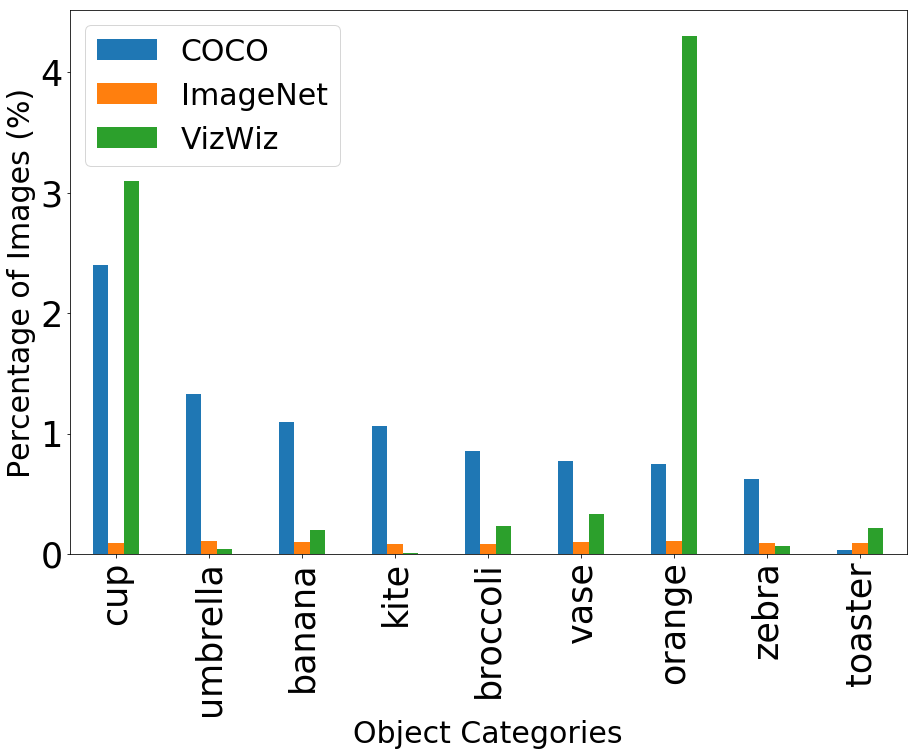}
    \caption{} \label{fig:2a}
    \end{subfigure}
    \hfill 
	\vspace{0.6em}
	\begin{subfigure}{0.32\textwidth}
    \includegraphics[width=\linewidth]{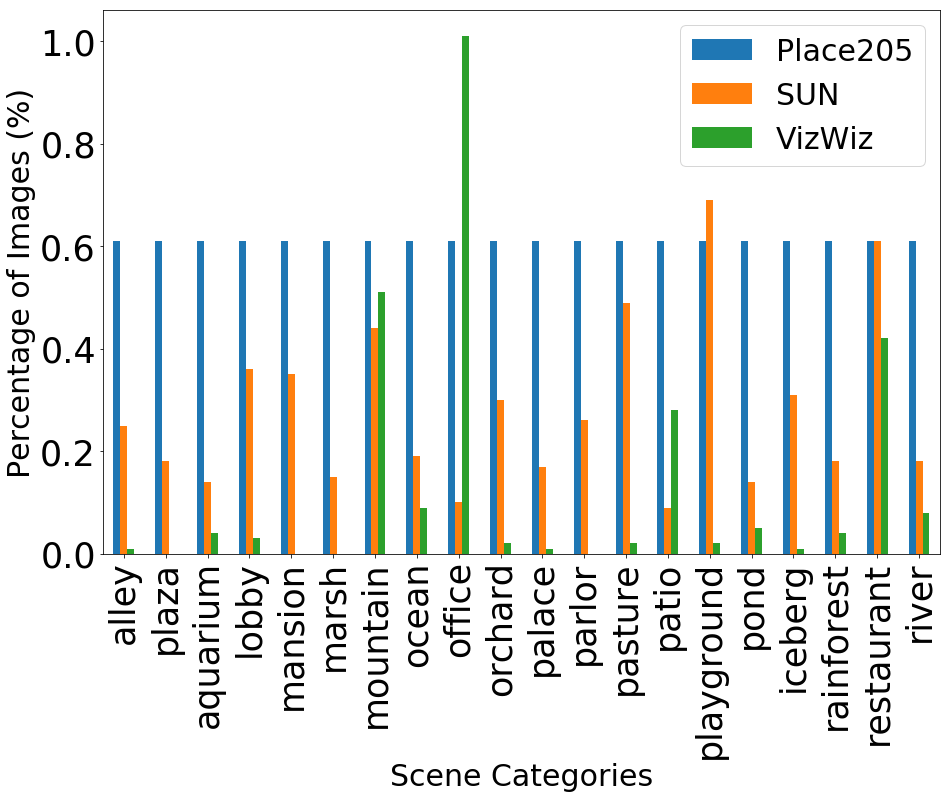}
    \caption{} \label{fig:2b}
    \end{subfigure}
    \hfill 
    \vspace{0.6em}
    \begin{subfigure}{0.32\textwidth}
    \includegraphics[width=\linewidth]{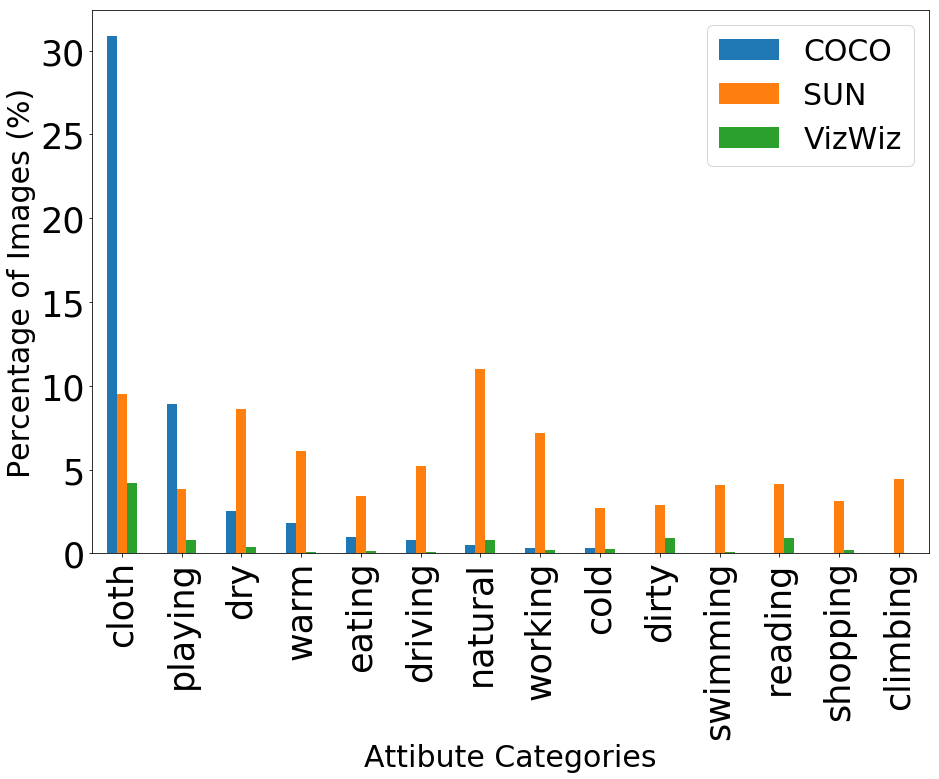}
    \caption{} \label{fig:2c}
    \end{subfigure}
    \hfill
    \vspace{-1.5em}
  \caption {Parallel results to those in Figure 2 of the main paper, showing the fraction of all images in our VizWiz-Captions and popular vision datasets that contain each category for the following vision problems: (a) object recognition, (b) scene classification, and (c) attribute recognition.}
  \label{fig_relativeImageClassificationComparison}
\end{figure*}

\begin{figure}[h!]
\centering
\includegraphics[width=\textwidth]{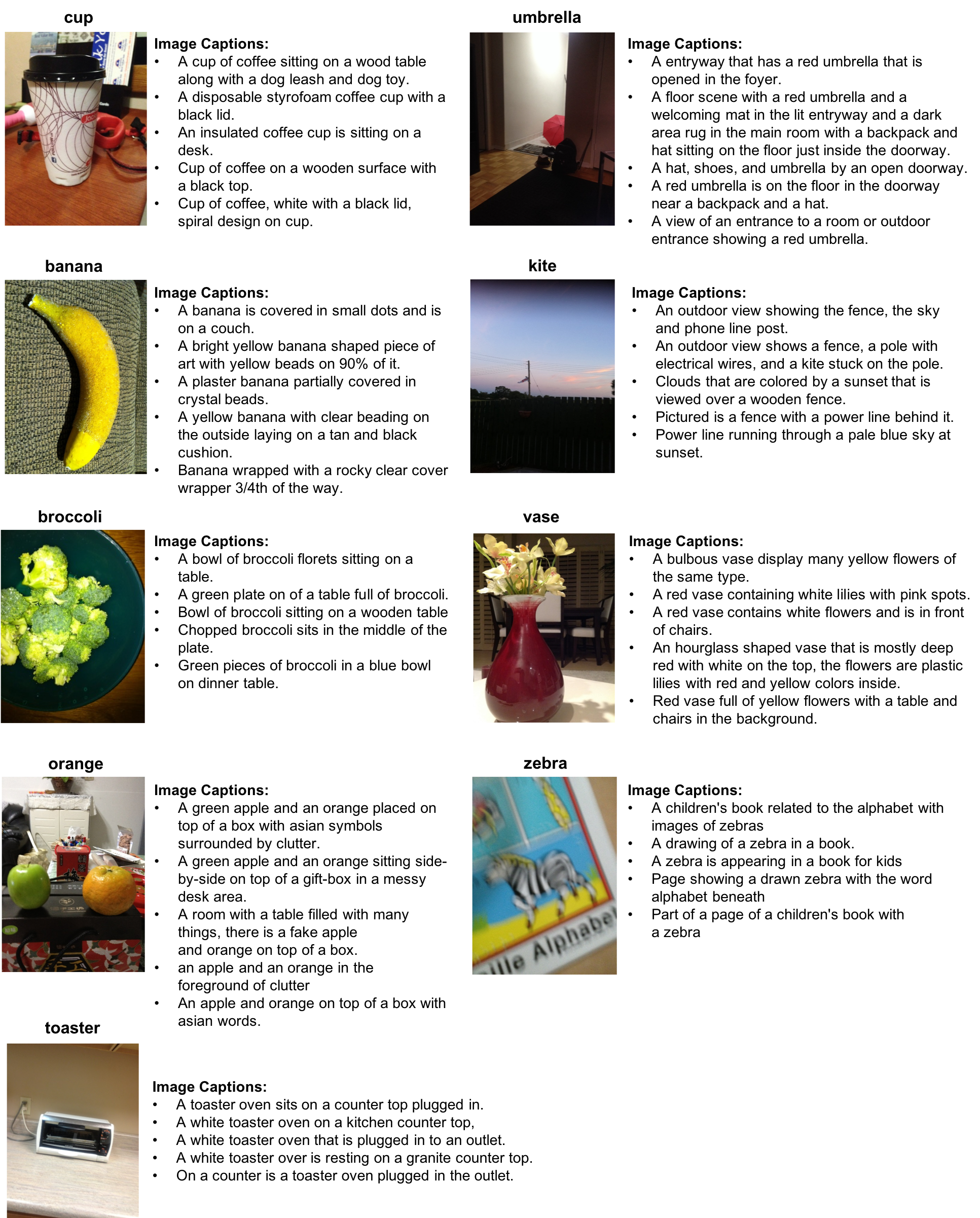}
\caption{Examples of images showing visual concepts that are common in existing computer vision datasets for the object recognition task.}
\label{fig_exampleObjectRecognitionImages}
\end{figure}

\begin{figure}[h!]
\centering
\includegraphics[width=\textwidth]{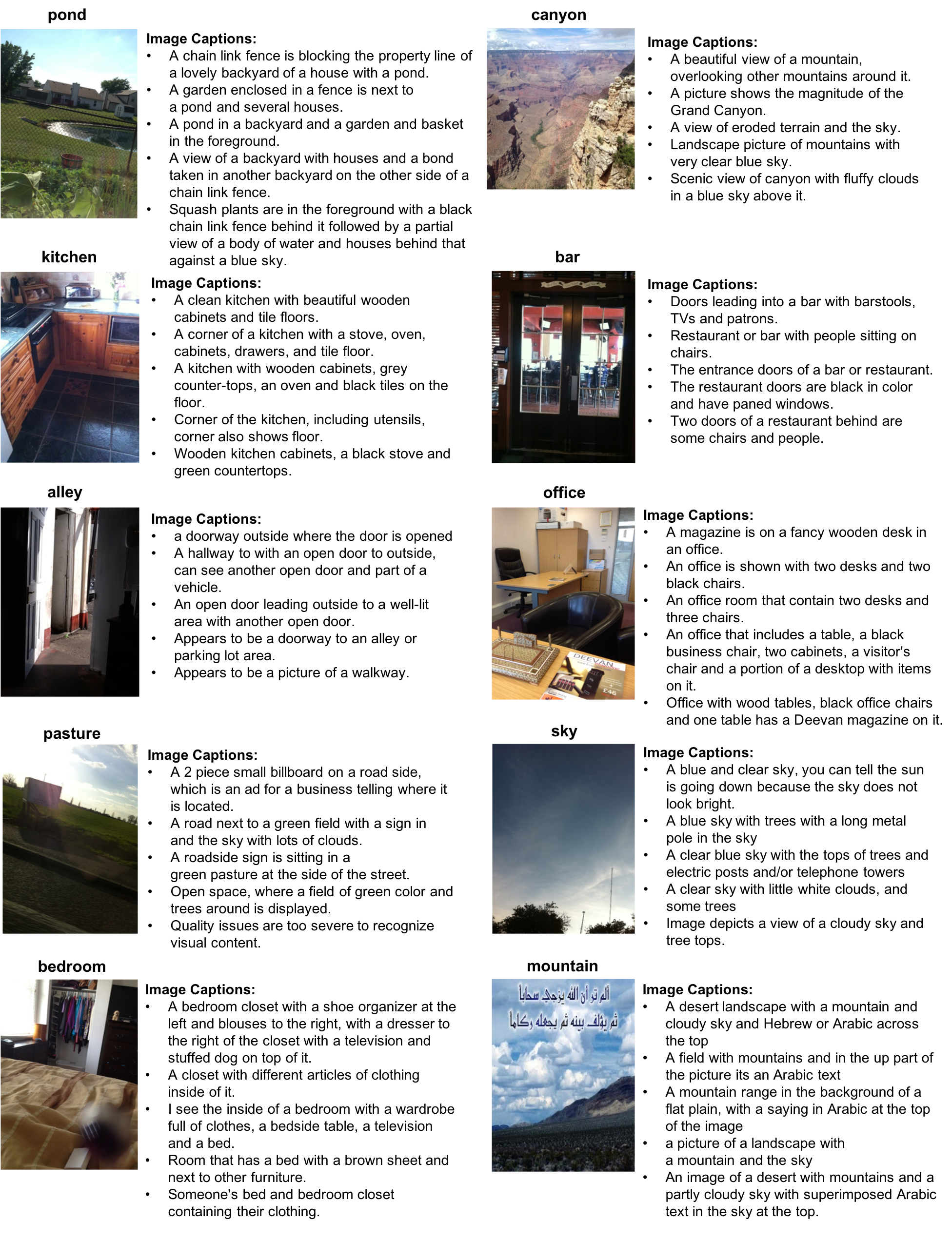}
\caption{Examples of images showing visual concepts that are common in existing computer vision datasets for the scene recognition task.}
\label{fig_exampleSceneRecognitionImages}
\end{figure}

\begin{figure}[h!]
\centering
\includegraphics[width=\textwidth]{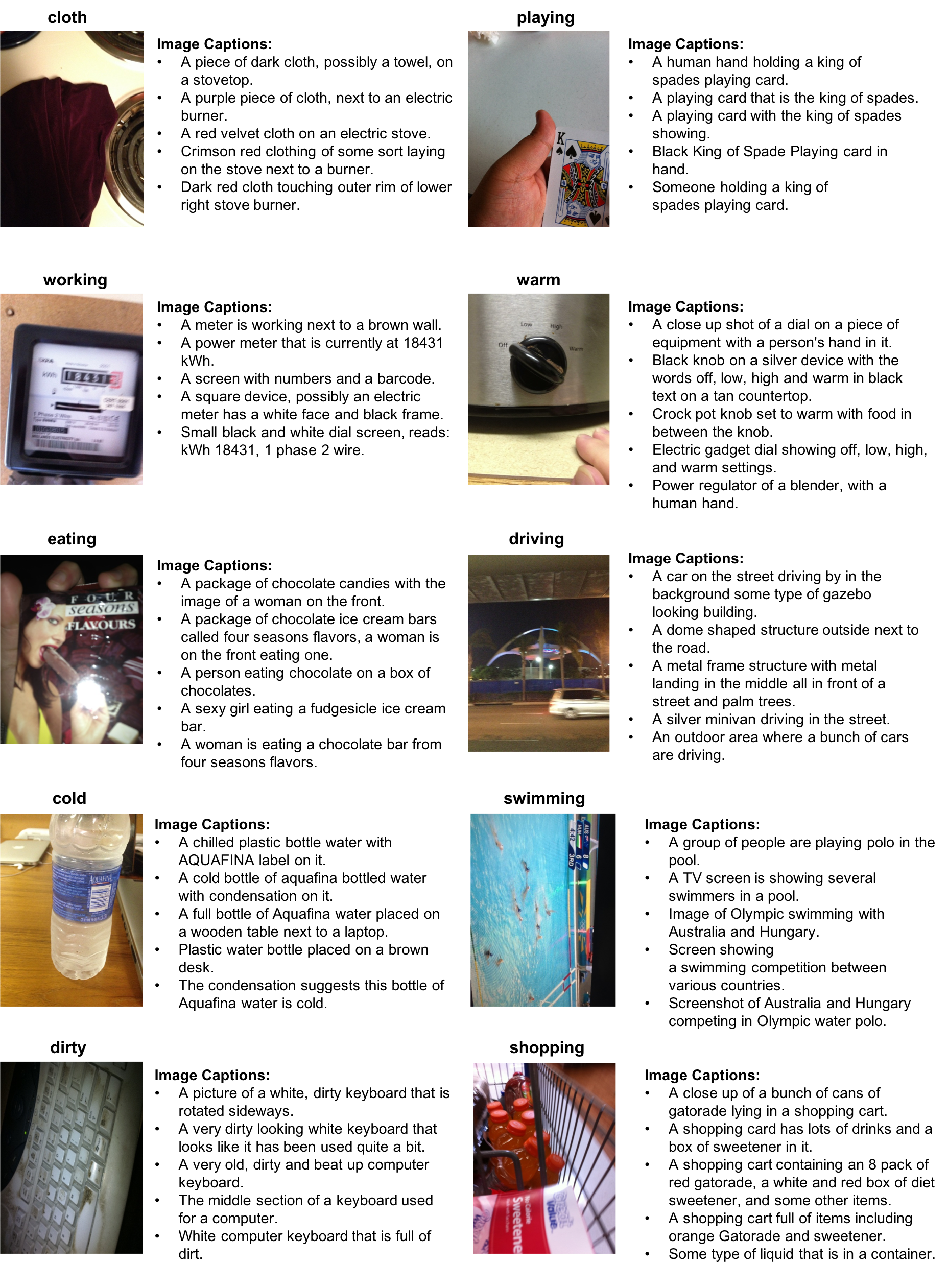}
\caption{Examples of images showing visual concepts that are common in existing computer vision datasets for the attribute classification task.}
\label{fig_exampleAttributeRecognitionImages}
\end{figure}

\clearpage
Finally, we provide histograms showing the relative prevalence of categories found in the mainstream computer vision datasets versus our dataset for three image classification tasks: recognizing objects, scenes, and attributes.  Results are shown in Figure~\ref{fig_relativeImageClassificationComparison}, complementing those shown in the main paper in  Figure 2.  Exemplar images in our dataset that show concepts overlapping with those in the mainstream computer vision datasets are shown in Figures~\ref{fig_exampleObjectRecognitionImages}, \ref{fig_exampleSceneRecognitionImages}, and \ref{fig_exampleAttributeRecognitionImages}.

%% file: supp-algorithm-benchmarking.tex
\clearpage
\section{Algorithm Performance with Data Augmentation (supplements Section 4)}
\label{sec_dataAugmentation}
Given the need for improved algorithm performance for low quality images, we examined the potential for using data augmentation during training to help the models better cope with low quality images at test time.  For this analysis, we chose the AoANet algorithm since it has the best performance when training from scratch.  We re-trained this algorithm from scratch again using VizWiz-Captions, but this time augmented a copy of all the training images after blurring them using a 15x15 averaging filter kernel on all the training images. 

Results are shown in Table~\ref{table_algBenchmarkingDataAugmentation}.  We observe that the performance is worse with data augmentation; e.g., with respect to the CIDEr score, overall performance falls from 60.5 to 56.2.  We suspect this performance drop is because artificial image distortions are unsuitable for mimicking real-world quality issues~\cite{chiu2020assessing}, and thus distract from model training.

\begin{table*}[h!]
  \centering
        \begin{tabular}{ c  c  c  c  c  c  c  c  c  c}
    
    \toprule
       && \bf B@1 & \bf B@2 & \bf B@3 & \bf B@4 & \bf METEOR & \bf ROUGE & \bf CIDEr & \bf SPICE  \\
     \midrule
     \multirow{5}{*}{{{\bf AOANet~}}}  
        & All & 66.4 & 47.0 & 32.3 & 22.1 & 20.0 & 46.5 & 56.2 & 13.8  \\
        \cdashline{2-10}
        & Easy & 69.7 & 50.6 & 35.5 & 24.6 & 21.2 & 49.1 & 60.3 & 14.2 \\
        & Medium & 63.3 & 42.7 & 27.7 & 18.2 & 18.5 & 43.2 & 50.8 & 13.6  \\
        & Difficult & 36.3 & 19.8 & 11.0 & 6.4 & 12.3 & 29.7 & 40.3 & 11.3 \\
    \bottomrule
  \end{tabular}
        \caption{Analysis of the top-performing image captioning algorithm when trained from scratch using data augmentation of blurred images.  Results are shown on all test images as well as only the subsets deemed easy, medium and difficult. (B@ = BLEU-)}
        ~\label{table_algBenchmarkingDataAugmentation}
        \vspace{-2.5em}
\end{table*} 

We also report human performance based on the same set of evaluation metrics.  To do so, we only consider images with five valid captions (i.e., ``easy" images).  We randomly choose one caption per image as the prediction, and use the remaining four captions for evaluation.  Results are shown in Table~\ref{table_algBenchmarkingDataAugmentation}.

\begin{table*}[h!]
  \centering
        \begin{tabular}{ c  c  c  c  c  c  c  c  c  c}
    \toprule
       && \bf B@1 & \bf B@2 & \bf B@3 & \bf B@4 & \bf METEOR & \bf ROUGE & \bf CIDEr & \bf SPICE  \\
     \midrule
        & Easy & 60.3 & 40.7 & 27.7 & 18.8 & 22.0 & 43.4 & 83.5 & 17.5  \\
    \bottomrule
  \end{tabular}
        \caption{Human performance on all test images deemed easy. (B@ = BLEU-)}
        \label{table_algBenchmarkingDataAugmentation}
\end{table*} 
\vspace{-3em}